\documentclass[letterpaper]{article} 
\usepackage{aaai2026}  
\usepackage{times}  
\usepackage{helvet}  
\usepackage{courier}  
\usepackage[hyphens]{url}  
\usepackage{graphicx} 
\usepackage{natbib}  
\usepackage{caption} 
\usepackage{algorithm}
\usepackage{algorithmic}
\usepackage{style}

\usepackage{newfloat}
\usepackage{listings}
\DeclareCaptionStyle{ruled}{labelfont=normalfont,labelsep=colon,strut=off} 
\floatstyle{ruled}
\newfloat{listing}{tb}{lst}{}
\floatname{listing}{Listing}
\nocopyright 

\title{Strategic Communication under Threat: Learning Information Trade-offs in Pursuit-Evasion Games}
\author {
    Valerio La Gatta\textsuperscript{\rm 1},
    Dolev Mutzari\textsuperscript{\rm 2},
    Sarit Kraus\textsuperscript{\rm 2},
    VS Subrahmanian\textsuperscript{\rm 1}
}
\affiliations {
    \textsuperscript{\rm 1}Department of Computer Science \& Buffett Institute for Global Affairs, Northwestern University\\
    \textsuperscript{\rm 2}Department of Computer Science, Bar Ilan University\\
}

\usepackage{bibentry}

\begin{document}

\maketitle

\begin{abstract}
Adversarial environments require agents to navigate a key strategic trade-off: acquiring information enhances situational awareness, but may simultaneously expose them to threats. To investigate this tension, we formulate a Pursuit-Evasion-Exposure-Concealment Game (PEEC) in which a pursuer agent must decide when to communicate in order to obtain the evader’s position. Each communication reveals the pursuer’s location, increasing the risk of being targeted. Both agents learn their movement policies via reinforcement learning, while the pursuer additionally learns a communication policy that balances observability and risk. We propose SHADOW (Strategic-communication Hybrid Action Decision-making under partial Observation for Warfare), a multi-headed sequential reinforcement learning framework that integrates continuous navigation control, discrete communication actions, and opponent modeling for behavior prediction. Empirical evaluations show that SHADOW pursuers achieve higher success rates than six competitive baselines. Our ablation study confirms that temporal sequence modeling and opponent modeling are critical for effective decision-making. Finally, our sensitivity analysis reveals that the learned policies generalize well across varying communication risks and physical asymmetries between agents.
\end{abstract}


\section{Introduction}

Intelligent agents operating in adversarial or high-stakes environments such as surveillance, search-and-rescue, or contested terrain, must often manage a fundamental strategic tension: the need to gather information for situational awareness versus the risk of being exposed~\cite{chen2015multi,liu2024multi}. This dilemma arises in many real-world scenarios, where communication and sensing actions not only provide critical data about an adversary’s position or intent, but also reveal the agent’s own presence or location to hostile observers. Decision-making systems that can reason about this trade-off are essential for enabling safe and effective autonomous behavior.

We address this challenge by introducing a Pursuit-Evasion-Exposure-Concealment Game (PEEC) where a pursuer seeks to intercept an evader under partial observability. The pursuer can choose to obtain the evader’s position, but doing so reveals its own location, potentially aiding the evader’s escape or increasing the risk of being eliminated. This PEEC formalizes the dilemma of acting to reduce uncertainty versus remaining covert to reduce risk. The game ends under one of the following conditions: (i) the pursuer captures the evader, i.e., their distance falls below a fixed capture radius, (ii) the pursuer is eliminated, i.e.,  it is shot with a certain probability when it chooses to query the evader’s position, or (iii) the evader escapes, i.e., a fixed time horizon is reached. 

Prior work in traditional Pursuit-Evasion Games (PEGs) has addressed partial observability \cite{rhodes1969differential, bernhard1988saddle} and cost-sensitive communication \cite{gupta2014common, aleem2015self, maity2023efficient, maity2024optimal, aggarwal2024linear}, but has rarely considered implicit exposure costs where the act of gathering information can itself be exploited. Our PEEC setup targets this trade-off: silence acts as a protective measure, while communication carries the risk of revealing the pursuer’s position. To our knowledge, the only prior work that explicitly models a PEEC setting is \cite{huang2021pursuit}, which offers a closed-form solution but makes strong simplifying assumptions, including symmetric agent goals (i.e., zero-sum game), guaranteed pursuer survivability (i.e., the pursuer cannot be eliminated when discovered), and favorable dynamics (e.g., higher maneuverability for the pursuer).

We relax these assumptions and propose \textbf{SHADOW} (Strategic-communication Hybrid Action Decision-making under partial Observation for Wargaming), a reinforcement learning (RL) framework for solving PEECs under realistic asymmetries and nonlinear dynamics\footnote{RL-based solutions exist for partially observable PEGs with multi-agent coordination \cite{de2021decentralized} and delayed communication \cite{wei2023differential, hu2024transfer, wang2025hierarchical}, but ignore the strategic cost of exposure. To our knowledge, no prior RL method explicitly targets a PEEC game.}. SHADOW learns both a continuous navigation policy and a discrete communication policy, jointly optimized to balance the benefit of acquiring information with the risk of exposure. Crucially, SHADOW agents also include an RL-based opponent modeling predictor to estimate the position of the adversary when the pursuer is not querying the evader's state. 

We instantiate a SHADOW-controlled pursuer and evader in a PEEC game. The pursuer strategically queries the evader’s position at the cost of being revealed, while the evader learns to evade under uncertainty (without communication). Our results show that SHADOW pursuers outperform both static and RL baselines by achieving higher success rates with fewer episodes and reduced exposure. SHADOW agents adapt their strategies to varying threat levels, communication costs, and speed disadvantages, and reduce unnecessary communication over time through opponent modeling. See the Appendix for illustrative examples of learned strategies and behaviors.

\noindent \textbf{Our Contribution}
    
\begin{enumerate}
    \item {\bf Generalization of PEECs}: We extend PEECs to accommodate non-holonomic and nonlinear dynamics, as well as asymmetric, non-quadratic payoffs.
    \item {\bf SHADOW, an RL Framework for PEECs}: We develop a corresponding RL model, designed to address this expanded class of PEECs. SHADOW employs dynamic opponent modeling to balance information acquisition with the risk of adversarial exposure.
    \item {\bf Cost of Information Acquisition}: We provide the first formal \emph{quantitative definition} of the cost of information acquisition in PEECs. It captures how much the pursuer is willing to pay per query under equilibrium behavior. Assuming zero-sum, we prove a non-negative lower bound, that is used in the experimental section. 
    \item {\bf Extensive Experimental Evaluation}: We systematically evaluate SHADOW across varied configurations, analyze learning dynamics, communication strategies, and performance under varying threat levels and agent speeds\footnote{Code will be made publicly available upon publication.}. Our results show that SHADOW pursuers learn to adapt their communication frequency, balance risk and reward, and outperform both periodic and RL-based baselines in pursuit success and efficiency.
 
\end{enumerate}

\section{Related Work}


    
    


Pursuit-Evasion Differential Games (PEGs) have long explored how agents operate under uncertainty, particularly in adversarial settings. Prior work models limited observability through three main approaches: (i) \emph{exogenous visibility limits} due to environmental constraints, (ii) \emph{internal sensing costs} that penalize information queries; and (iii) \emph{implicit exposure costs}, where observing reveals the agent’s own state to the opponent. While the first two have been extensively studied, implicit exposure remains underexplored. 
A full survey and comparison of these models is in the Appendix, where we review both classical and modern RL-based PEG formulations. Here, we focus on the third setting as it is both underexplored and central to our work. To our knowledge, the only prior work that explicitly models \emph{implicit exposure cost} is the PEEC framework in \cite{huang2021pursuit}, where information acquisition comes at the strategic cost of revealing one’s own state.\footnote{The notion that silence can itself be informative is also explored in~\cite{maity2023efficient}.} Specifically, \cite{huang2021pursuit} formalizes this trade-off within a two-player Linear Quadratic Gaussian (LQG) differential game, where each observation incurs both an explicit sensing cost and an implicit exposure cost—since querying the opponent’s state simultaneously discloses the querying agent’s position. The study decouples control and sensing decisions, proves the existence of Nash equilibria, and derives a closed-form solution characterized by a periodic ``sense–then–hide" policy. While this framework provides a foundational treatment of the exposure–information dilemma, it comes with notable limitations: (\romannumeral 1) it is restricted to the LQG setting, which enables tractability but lacks generality, (\romannumeral 2) it assumes  a zero-sum game, where agent incentives are strictly opposed, and (\romannumeral 3) it treats exposure as a purely strategic cost without modeling the physical risk of elimination. 


\section{Methodology}
We begin by outlining the key modeling challenges posed by PEEC games. We then present SHADOW, our learning architecture, and describe how its design addresses these challenges. 

\subsection{Key Modeling Challenges (CGs)}
\begin{enumerate}[label=CG\textsubscript{\arabic*}, ref=CG\textsubscript{\arabic*}]
    \item \label{challenge:variable-observation}
    \emph{Variable Observation Space}: The observation dimension received from the environment depends on the query action. Without querying, agents perceive only their local state. When queried, both agents receive the full environment state, complicating learning and representation
\item \label{challenge:temporal-dependence}
    \emph{Non-Markovian Dependencies}: Agents must condition their decisions on past observations and elapsed time. Policies must integrate temporal information and memory to handle delayed effects and shifting strategies.
\item \label{challenge:parameter-generalization}
    \emph{Parameter Generalization}: The environment is governed by parameters like speed, bounds on acceleration, and capture radius. Policies must adapt across parameter settings without retraining, enabling robust adaptation to new scenarios.
\item \label{challenge:hybrid-action}
    \emph{Hybrid Action Space}: The agents need to simultaneously execute both a continuous navigation action and a discrete query decision. 
\item \label{challenge:strategic-querying}
    \emph{Strategic Querying}: As for the PEEC formulation, querying the opponent’s state is beneficial but risky, as it reveals its own position. The agent must learn to query selectively, weighing informational gain against the cost of exposure.
\end{enumerate}

\subsection{SHADOW: Bird's-Eye View} \label{sec:method}
\begin{figure}[t]
    \centering
    \includegraphics[width=\linewidth]{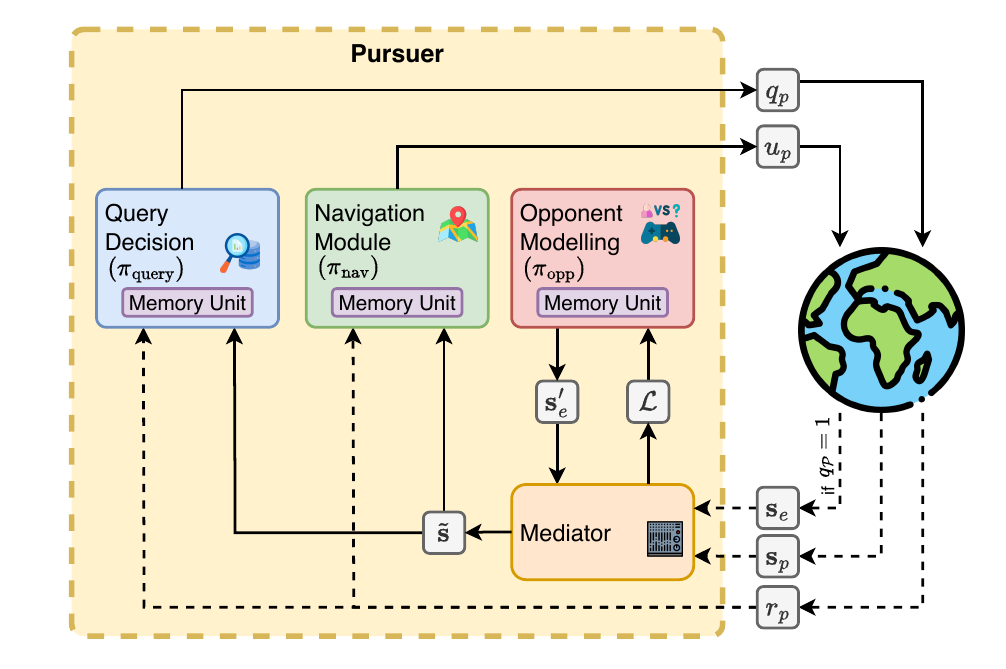}
    \caption{\textbf{SHADOW Pursuer}: The Pursuer operates its navigation control $u_p$ and decides whether to query the opponent's state via a binary action $q_p \in \{0,1\}$. The environment returns the updated pursuer state $\ss_p$ and, if $q_p = 1$, the evader’s current position $\ss_e$. The \emph{Mediator} determines the pursuer’s internal state representation $\tilde{\mathbf{s}}$, comprising: (i) the current position of the pursuer $\ss_p$, (ii) the elapsed time since the last observation, (iii) the last observed position of the evader, and (iv) the estimated current position of the evader which is either returned by the environment ($\ss_e$), if $q_p = 1$, or inferred by the \emph{Opponent Modeling} module ($\mathbf{s}_e'$) if $q_p = 0$. The \emph{Mediator} also provides feedback $\mathcal{L}$ to the \emph{Opponent Modeling} module, indicating prediction error when the true position of the adversary becomes available ($q_p = 1$). Finally, the pursuer's internal state $\tilde{\mathbf{s}}$ and reward $r_p$ are passed to the \emph{Query Decision} and \emph{Navigation Module} to decide next actions. All networks include a \emph{Memory Unit} (e.g., LSTM) responsible for encoding the temporal observation history.}
    \label{fig:model}
\end{figure}

Deriving a closed-form solution to our PEEC game is challenging. We therefore design SHADOW, an RL-based method to learn policies of the players. Figure~\ref{fig:model} shows the architecture of a SHADOW pursuer. The \emph{Navigation Module} determines continuous navigation control input $u_p$ at each timestep. The \emph{Query Decision Module} decides whether to query $q_p$ the evader’s current position, trading off information gain against potential risk of being discovered or eliminated. The \emph{Opponent Modeling Module} predicts the evader’s position $\ss_e'$ when no query is made, and is updated via $\mathcal{L}$ when ground-truth observations are available. Each module contains a recurrent \emph{Memory Unit} (e.g., LSTM) to capture temporal dependencies. The \emph{Mediator} integrates all available information (past positions, query outcomes, and timing) into a compact internal state $\tilde{\ss}$ that serves as input to both decision modules.

Due to the asymmetric configuration of our PEEC, a SHADOW evader shares the same architecture as the pursuer, except it lacks the \emph{Query Decision Module} as only the pursuer can access the adversary's position.

We now describe each component in greater detail. While the following discussion focuses on the pursuer, the same principles apply to the evader when equipped with a full SHADOW architecture.

\noindent \textbf {Mediator.}
\label{sec:state-representation}
The \emph{Mediator} translates raw observations from the environment into an internal representation for the agents, addressing \ref{challenge:variable-observation}. The observation of a pursuer $\mathcal{P}$ facing an evader $\Evader$ includes: (i) $\mathcal{P}$'s current position $\ss_p(t)$, (ii) the elapsed time since the last observation $(t-t_0)$, (iii)  $\Evader$'s last observed position $\ss_e(t_0)$, and (iv) the estimated current position of the evader $\ss_e'(t)$ which is inferred by the \emph{Opponent Modeling} module. When the pursuer queries the evader's state, the estimated and observed positions of the opponent coincide. This formulation ensures a consistent structure in the agent's observations, regardless of the pursuer’s query decision, while allowing both the pursuer and evader to implicitly assess the reliability of the estimated opponent's state.

In addition to positional information, the \emph{Mediator} incorporates key environmental parameters (e.g., the agents’ velocities, the capture radius, and shooting radius) into the internal state $\tilde{\ss}$. While these elements assume some prior knowledge of the opponent's capabilities, they also enable agents to generalize across varying scenarios, supporting adaptive policy learning and addressing \ref{challenge:parameter-generalization}.

\noindent \textbf{Navigation \& Query Decision.}
The pursuer operates in a hybrid action space involving two components: a binary decision $q_p(t) \in \{0,1\}$ determining whether to query the evader's position at time $t$, and a continuous decision $u_p(t)$ controlling its navigation policy. To address this, we use a decoupled learning framework, where separate RL agents are responsible for each decision.

Following~\cite{wei2023differential}, we learn the navigation policy $\pi_{\text{nav}}$ using the Twin Delayed Deep Deterministic Policy Gradient (TD3) algorithm~\cite{fujimoto2018addressing}, which is well suited for continuous control tasks. Specifically, the pursuer $\Pursuer$ receives its internal state observation $\tilde{\ss}$ (processed via the \emph{Mediator}) and outputs an action $u_p = \pi_{\text{nav}}(\tilde{\ss})$ to control movement dynamics.

Simultaneously, the pursuer’s query policy $\pi_{\text{query}}$ is trained using Proximal Policy Optimization (PPO)~\cite{schulman2017proximal}, a policy-gradient algorithm robust to stochastic discrete actions. At each timestep, $\Pursuer$ takes a query decision $q_p = \pi_{\text{query}}(\tilde{\ss})$, determining whether to access the opponent's real position. 

Both the TD3 and PPO policies leverage a sequence model, specifically an LSTM layer, which acts as a memory unit to encode the temporal observation history. This design addresses challenge~\ref{challenge:temporal-dependence}, enabling policies to learn from both current and past information\footnote{While we include the elapsed time since last communication in the state of the agents, the LSTM-based memory may also represent previous observations and their effect on agent dynamics.}.

This modular design where the navigation and query policies are learned independently provides flexibility w.r.t. \ref{challenge:hybrid-action}, and facilitates targeted policy optimization. We compare this design against monolithic baselines that handle hybrid action spaces jointly in the experiments. 

\noindent \textbf{Opponent Modeling.}
The opponent modeling module estimates the position of the opponent and quantifies the associated uncertainty. This information can support both navigation ($\pi_\text{nav}$) and query ($\pi_\text{query}$) decision-making: accurate predictions may enable more effective maneuvering (for both the pursuer and the evader) when direct observations are unavailable, and reduce the need to query when there is confidence in the opponent’s estimated position, thus addressing challenge \ref{challenge:strategic-querying}.

We model this component as a TD3 agent which predicts the evader's position $\ss_e'(t)$ and related uncertainty, from $\Evader$'s last observed state $\mathbf{s}_e(t_0)$ and the time since last observation $t-t_0$. Formally, the agent learns the following policy:
\begin{equation*}
    \pi_{\text{opp}}: (\mathbf{s}_e(t_0), t-t_0) \rightarrow ({\ss}_e'(t), \sigma)
\end{equation*}
where $\sigma \in \mathbb{R}$ is the predicted uncertainty. The model minimizes a Gaussian Negative Log-Likelihood (NLL) loss:
\begin{equation*}
\mathcal{L} = \frac{|\mathbf{s}_e(t) - \mathbf{s}_e'(t)|}{2\sigma + \varepsilon} + \frac{1}{2} \log(\sigma + \varepsilon),
\end{equation*}
where the first term penalizes inaccurate predictions (scaled by uncertainty), while the second prevents trivial solutions with overly large $\sigma$. A small constant $\varepsilon$ ensures numerical stability.  The predicted uncertainty $\sigma$ quantifies reliability and can modulate decisions in $\pi_\text{nav}$ and $\pi_\text{query}$. Its effects and relationship with the elapsed time since the last communication ($t-t_0$) are examined in the experiments.

Since both $\Pursuer$ and $\Evader$ adapt their strategies during training (potentially in response to the predictions of the opponent model), it is essential to co-train $\pi_{\text{opp}}$ jointly with the navigation ($\pi_{\text{nav}}$) and query decision ($\pi_{\text{query}}$) policies. This ensures mutual adaptation and prevents policy misalignment due to static or outdated opponent predictions, justifying the use of an RL-based agent over a fixed pre-trained model for prediction of opponent state.

\section{A Concrete PEEC Game}
\label{sec:game}

To evaluate our learning framework, we instantiate a concrete PEEC game in a two-dimensional environment. The game involves a pursuer $\Pursuer$ and an evader $\Evader$ interacting on a bounded planar map $\Map \subset \RR^2$. While the evader moves covertly, the pursuer can choose to query the evader’s full state $\ss_e$ at the expense of revealing its own state $\ss_p$.

\noindent \textbf{State.} The state of the game $\ss=(\ss_p,\ss_e)$ consists of the local state of each player. The state of each player $i \in \{\Pursuer, \Evader\}$ is defined as $\ss_i=(x_i,y_i,\psi_i) \in \Map \times [-\pi, \pi)$ and includes their location $(x_i,y_i)$ and heading angle $\psi_i$. 

\noindent \textbf{Dynamics \& Actions.} Following~\cite{kothari2014cooperative}, the dynamics of agent $i\in \{\Pursuer,\Evader\}$ are given by
\begin{align*}
    \label{eq:dynamics}
    \dot{x}_i = v_i \cos{\psi_i},~~~~
    \dot{y}_i = v_i \sin{\psi_i},~~~~
    \dot{\psi}_i &= u_i / v_i
\end{align*}
The pair $(x_i(t),y_i(t)) \in \Map$ denotes the position of player $i$ at time $t$, $\psi_i(t)$ is its heading, $v_i$ is the constant velocity of player $i$, and $u_i(t) \in [-U_i, +U_i]$ is its lateral acceleration, which acts as a control input.

\noindent \textbf{Querying \& Observability.} In addition to controlling its lateral acceleration, the pursuer can \emph{query} the evader's state by contacting its control unit. We denote by $q_p(t) \in \{0,1\}$ the binary control variable for querying at time $t$. When $q_p(t) = 1$, both $\Pursuer$ and $\Evader$ observe the full state $\ss(t)$; otherwise, the agents only retain their local information $\ss_i(t)$. Thus, the game is partially observable, with observations determined by the pursuer's querying policy. 

\noindent \textbf{Game Evolution and Terminal Condition.} The game starts at time $t = 0$ from an initial state $\ss(0)$. Agents continuously evolve their trajectories by selecting $u_p(t)$ and $u_e(t)$, and the pursuer optionally issues queries via $q_p(t)$. The game terminates at the earliest time $\Horizon_f \le \Horizon$ when one of the following conditions is met: (\romannumeral 1) The pursuer catches the evader, i.e., the Euclidean distance $r(t)=d(\Pursuer,\Evader)$ falls below a capture threshold $r_\catch$. (\romannumeral 2) The evader survives until the terminal time $t = \Horizon$. (\romannumeral 3)  The pursuer communicates ($q_p(t)=1$) and is eliminated with probability $p_\eliminate = 2^{-r(t)/r_\eliminate}$, where $r_\eliminate$ is the shooting radius\footnote{Shooting is not modeled as a strategic decision of the evader. We assume that the evader shoots anytime the pursuer reveals its position, but it might miss the target depending on their distance.}.


\noindent \textbf{Pursuer Payoff Function.} The pursuer's payoff function $P_p$ includes an integral cost over time and a terminal reward:

\begin{equation*}
P_p = R_p^f - \int_{0}^{\Horizon_f}{(\alpha^\Time_p \mathbbm{1} + \alpha^\Query_p \mathbbm{1}_{q_{\Pursuer}} + \alpha^\Boundary_p \mathbbm{1}_{\ss_p \cap \partial \Map} + \alpha^\Acceleration_p |u_p|) dt} 
\end{equation*}

Here, $\alpha^\Time_p, \alpha^\Query_p, \alpha^\Boundary_p, \alpha^\Acceleration_p \ge 0$ are fixed coefficients that determine the cost profile:  

\begin{itemize}
    \item The \emph{time penalty} $\alpha^\Time_p$ encourages faster pursuit~\cite{jabeur2025robotic}.
    \item The \emph{query penalty} $\alpha^\Query_p$ reflects the cost or risk associated with revealing the pursuer's position.
    \item The \emph{boundary penalty} $\alpha^\Boundary_p$ penalizes collisions with the map boundary $\partial \Map$, causing physical damage to the UAV.
    \item The \emph{acceleration penalty} $\alpha^\Acceleration_p$ models energy or resource consumption due to lateral control effort.
    
\end{itemize}

The \emph{terminal reward} $R_p^f$ is given by:

    \begin{equation*}
        R_p^f(\ss(\Horizon_f)) =
        \begin{cases}
        r_p, & \Pursuer ~~\text{catches}~~ \Evader \\
        0, & \Horizon_f=\Horizon \\
        -p_p, & \Pursuer ~~\text{is eliminated}
        \end{cases}
    \end{equation*}

\noindent \textbf{Evader Payoff Function.} The evader integral payoff function takes a similar form, except $\alpha^\Query_e=0$ as $\Evader$ cannot query the state, and $\alpha^\Time_e = - \alpha^\Time_p \le 0$ to promote evasion. In addition, the evader's terminal reward $R_{\mathcal{E}}^f(\mathbf{s}_f)$ is  negative, i.e., $r_e=-r_p$ in case it gets caught, and zero otherwise. Since shooting is not modeled as a strategic decision, we do not reward the evader when the pursuer is eliminated.

\noindent \textbf{Nash Equilibrium (NE).} A pair of control strategies $\langle (u_p^*, q_p^*); u_e^* \rangle$ is an NE if players' payoffs are minimized,
\begin{align*}
(u_p^*, q_p^*) &\in \arg\min\limits_{(u_p, q_p)} {P_p(\langle (u_p, q_p) , u_e^* \rangle; \ss_0)} \\
u_e^* &\in \arg\min\limits_{u_e} {P_e(\langle (u_p^*, q_p^*) , u_e \rangle; \ss_0)}
\end{align*}

We denote the set of all NE solutions by $\Omega_\NE$. It is important to note that we do not assume a system-level payoff $P_S := P_p - P_e$ that one player minimizes and the other maximizes, as in zero-sum settings~\cite{ni2018improved,wei2023differential}. While such games admit elegant minimax solutions, they rely on strong assumptions about goal alignment. In our PEEC game, each agent optimizes its own payoff, reflecting potentially conflicting objectives. Even under simplified assumptions, we are not aware of a closed-form NE for the proposed game.


Next, we propose a formal definition for the non-monetary cost of information acquisition in PEEC games.

\begin{Definition}[Critical Information Acquisition Cost ($\CIAC$)]
The \emph{Critical Information Acquisition Cost ($\CIAC$)} is the threshold communication penalty $\alpha^\Query_c$, for which there exists an NE where the pursuer obtains a non-negative payoff:
$$ \alpha_c^\Query = \sup \{\alpha^\Query_p \mid \max\limits_{\Omega_{\NE}[\alpha^\Query_p]} {\EE[P_p | \alpha^\Query_p] \ge 0}\} $$
\end{Definition}
By definition, when the communication penalty $\alpha^\Query_p$ exceeds $\alpha^\Query_c$, a pursuer facing a rational evader cannot afford to communicate while ensuring a positive payoff. Conversely, when $\alpha^\Query_p < \alpha^\Query_c$, there exist a non-trivial querying strategy that yields a positive payoff.

\begin{restatable}{Proposition}{CiacPositive}
\label{prop:ciac-exist-positive}
With a zero-sum assumption (i.e., $P_e \equiv -P_p$) and $r_e=0$, $\alpha_c^\Query \ge 0$ is a maximum.
\end{restatable}

Intuitively, since $\EE[P_p|\alpha^\Query_p]$ is linear in $\alpha^\Query_p$ and so monotonic and continuous, and $\EE[P_p|\alpha^\Query_p=0] > -\infty$, $\alpha_c^\Query$ exists. Furthermore, fixing $\alpha^\Query_p<0$, the pursuer may rapidly and repeatedly query the state sufficiently many times to ensure a positive payoff. In our experiments, we measure a lower bound signal for $\CIAC$, defined below:

\begin{Definition}[Base Information Acquisition Cost ($\BIAC$)]
Given an NE $\langle (u_p^0, q_p^0); u_e^0 \rangle \in \Omega_\NE[\alpha^\Query_p=0]$, the \emph{Base Information Acquisition Cost ($\BIAC$)} is the maximal penalty $\underline{\alpha}_c^\Query$ a pursuer is willing to pay per query while ensuring a non-negative payoff:
$$ \underline{\alpha}_c^\Query = \frac{\EE[P_p \mid \alpha_p^\Query = 0]}{\EE[N^\Query_p]},$$
where $N^\Query_p$ is the number of pursuer queries.
\end{Definition}

\begin{restatable}{Proposition}{BiacLB}
\label{prop:biac-lower-bound}
Assuming zero-sum (i.e., $P_e \equiv -P_p$), $\underline{\alpha}_c^\Query \le \alpha_c^\Query$.
\end{restatable}

Intuitively, since $\langle (u_p^0, q_p^0); u_e^0 \rangle$ is an NE, the evader has no incentive to deviate, and therefore as long as $\alpha^\Query_p < \underline{\alpha}_c^\Query$, the pursuer can ensure an expected positive payoff without changing its strategy. Therefore, $\underline{\alpha}_c^\Query \le \alpha_c^\Query$.

\emph{Formal proofs of Propositions~\ref{prop:ciac-exist-positive},\ref{prop:biac-lower-bound} are in the Appendix.}

\section{Experimental Results}

\subsection{Experimental Setup}

We instantiate our PEEC game using a SHADOW pursuer following the architecture described above. The \emph{Evader} is also SHADOW-operated but omits the \emph{Query Decision} model $\pi_\text{query}$. As our PEEC formulation is asymmetric, only the pursuer can query the full state of the game. 

For all experimental settings, models were trained for 20,000 episodes using a mini-batch size of 32, and evaluated on the same $N=500$ held-out episodes. Unless otherwise specified, we retained the default hyperparameters provided in the original implementations of each algorithm. Further details are in the appendix.

\subsection{Experimental Protocol}

To evaluate SHADOW's effectiveness in learning adaptive communication and movement strategies in a pursuit-evasion game, we designed three experimental tracks: \emph{Baseline Comparison}, \emph{Ablation Study}, and \emph{Training Dynamics Analysis}. 
A few illustrative examples of game trajectories are provided in the Appendix.

\noindent \textbf{Baseline Comparison} We examine whether SHADOW outperforms three heuristic approaches and three RL-based strategies: 
\emph{(i)} \emph{No Communication}: the pursuer never communicates.
\emph{(ii)}  \emph{Random Communication}: the pursuer uses the inverse probability of getting shot to decide when to query the evader's state, $p_{\text{comm}}=1-p_{\text{shot}}$. 
\emph{(iii)} \emph{Periodic Communication}: the pursuer communicates periodically, each $k$ timesteps. This strategy was proven to be theoretically optimal in the setting of~\cite{huang2022pursuitevasiondifferentialgamestrategic}\footnote{\cite{huang2022pursuitevasiondifferentialgamestrategic} proposed a simplified PEEC game with strong assumptions, including symmetric agent goals (i.e., zero-sum formulation), guaranteed pursuer survivability (i.e., the pursuer cannot be eliminated when discovered), and favorable dynamics (e.g., higher maneuverability for the pursuer).}.
\emph{(iv)}  \emph{MultiHead PPO} \cite{flet2019merl}: the pursuer leverages a multi-headed actor with PPO to jointly learn the communication and movement policies. 
\emph{(v)}  \emph{P-DQN} \cite{xiong2018parametrized}: the pursuer leverages a Parametrized Deep Q-Network to jointly learn the query and navigation policies. 
\emph{(vi)}  \emph{HyAR} \cite{DBLP:journals/corr/abs-2109-05490}: the pursuer learns the relationship between the discrete action ($q_p$) and the continuous action ($u_p$) using a variational autoencoder.

Performance is assessed via metrics in three categories: (\romannumeral 1) \emph{End-State Outcomes} includes the percentage of evaluation episodes the pursuer wins $P_{\text{win}}$, gets shot $P_{\text{shot}}$ or runs out of time $P_{\text{timeout}}$; (\romannumeral 2) \emph{Communication Strategy} includes the average percentage of communication events $C_{\text{ratio}}$, the average time between queries $C_{\text{gap}}$, the average distance between agents at the last communication $D_{\text{comm}}$, and $\BIAC$; and (\romannumeral 3) \emph{Behavioral Efficiency} includes the average episode duration $T_{\text{len}}$ and steering costs $\bar{S}_P,\bar{S}_E$.

For each metric, we also report the standard deviation across evaluation episodes. Full details on the baselines and metrics are provided in the Appendix.

\noindent \textbf{Ablation Study} We quantify the contribution of SHADOW’s internal components, focusing on the \emph{Opponent Modeling} and LSTM-based \emph{Memory Unit}. For the opponent modeling module, we evaluate four configurations, where both agents may or may not be equipped with opponent modeling.
Similarly, we consider both agents with and without the LSTM layer. This enables us to assess the extent to which temporal abstraction contributes to effective movement and whether the advantage of temporal memory depends on mutual availability.

\noindent \textbf{Sensitivity Analysis} We measure SHADOW's performance under varying environmental conditions, i.e., the shooting radius $r_\eliminate$ and the speed ratio $v_e/v_p$, and the uncertainty $\sigma$ predicted by the \emph{opponent modeling} module. This reveals how the pursuer adapts to changing communication risks and how physical capabilities constrain strategic flexibility.

\noindent \textbf{Training Dynamics}
We investigate how the pursuer and evader strategies evolve during training by tracking their metrics over time. Specifically, we analyze how the pursuer learns to balance its communication strategy, and reduce its movement cost and risk of being shot. 

\subsection{Results}

\begin{table*}[t]
\centering
\resizebox{\textwidth}{!}{
\begin{tabular}{r|ccc|ccc|ccc} \toprule
& \multicolumn{3}{c}{End-State Outcomes} & \multicolumn{3}{c}{Communication Strategy} & \multicolumn{3}{c}{Behavioral Efficiency} \\

Model                & $P_{\text{win}}$ & $P_{\text{shot}}$ & $P_{\text{timeout}}$ & $C_{\text{ratio}}$ & $C_{\text{gap}}$ & $C_{\text{comm}}$ & $T_{len}$ & $\bar{S}_P$ & $\bar{S}_E$  \\ \midrule
No communication     & $0.184 \pm 0.034$ & N/A               & $0.816 \pm 0.034$ & N/A               & N/A               & N/A               & $143.2 \pm 44.37$ & $0.192 \pm 0.003$ & $0.233 \pm 0.012$  \\
Random communication & $0.020 \pm 0.012$ & $0.980 \pm 0.012$ & $0.000 \pm 0.000$ & $0.492 \pm 0.014$ & $2.018 \pm 0.054$ & $0.194 \pm 0.012$ & $32.82 \pm 2.149$ & $0.149 \pm 0.013$ & $0.313 \pm 0.019$  \\ \midrule
Periodic  (k=5)      & $0.182 \pm 0.033$ & $0.818 \pm 0.033$ & $0.000 \pm 0.000$ & $0.204 \pm 0.003$ & $5$               & $0.149 \pm 0.010$ & $43.46 \pm 2.363$ & $0.126 \pm 0.009$ & $0.223 \pm 0.010$  \\
Periodic  (k=10)     & $0.264 \pm 0.038$ & $0.736 \pm 0.038$ & $0.000 \pm 0.000$ & $0.104 \pm 0.003$ & $10$              & $0.143 \pm 0.010$ & $57.05 \pm 2.819$ & $0.112 \pm 0.007$ & $0.198 \pm 0.011$  \\
Periodic  (k=20)     & $0.480 \pm 0.043$ & $0.520 \pm 0.043$ & $0.000 \pm 0.000$ & $0.056 \pm 0.003$ & $20$              & $0.159 \pm 0.011$ & $86.37 \pm 4.857$ & $0.125 \pm 0.006$ & $0.181 \pm 0.009$  \\
Periodic  (k=30)     & $0.546 \pm 0.041$ & $0.439 \pm 0.043$ & $0.015 \pm 0.001$ & $0.039 \pm 0.004$ & $30$              & $0.191 \pm 0.013$ & $119.2 \pm 6.361$ & $0.159 \pm 0.006$ & $0.191 \pm 0.009$  \\
Periodic  (k=40)     & $0.576 \pm 0.052$ & $0.416 \pm 0.036$ & $0.010 \pm 0.008$ & $0.033 \pm 0.004$ & $40$              & $0.206 \pm 0.012$ & $320.1 \pm 22.93$ & $0.168 \pm 0.005$ & $0.123 \pm 0.005$  \\
Periodic  (k=50)     & $0.276 \pm 0.039$ & $0.274 \pm 0.039$ & $0.450 \pm 0.043$ & $0.026 \pm 0.004$ & $50$              & $0.244 \pm 0.011$ & $283.3 \pm 29.01$ & $0.162 \pm 0.005$ & $0.121 \pm 0.005$  \\ \midrule
MultiHead PPO        & $0.272 \pm 0.039$ & $0.056 \pm 0.020$ & $0.672 \pm 0.041$ & $0.066 \pm 0.019$ & $10.32 \pm 0.020$ & $0.504 \pm 0.021$ & $258.5 \pm 39.85$ & $0.162 \pm 0.001$ & $0.185 \pm 0.012$  \\
P-DQN                & $0.396 \pm 0.043$ & $0.042 \pm 0.017$ & $0.564 \pm 0.043$ & $0.046 \pm 0.001$ & $62.13 \pm 5.546$ & $0.448 \pm 0.018$ & $441.1 \pm 37.59$ & $0.196 \pm 0.012$ & $0.244 \pm 0.013$  \\ 
HyAR                 & $0.246 \pm 0.037$ & $0.002 \pm 0.003$ & $0.752 \pm 0.037$ & $0.002 \pm 0.001$ & $55.01 \pm 648.1$ & $0.686 \pm 0.065$ & $52.94 \pm 8.399$ & $0.897 \pm 0.012$ & $0.238 \pm 0.011$ \\
\midrule
SHADOW (Ours)        & $0.620 \pm 0.042$ & $0.350 \pm 0.041$ & $0.030 \pm 0.015$ & $0.172 \pm 0.023$ & $29.42 \pm 4.988$ & $0.240 \pm 0.014$ & $272.8 \pm 20.04$ & $0.168 \pm 0.011$ & $0.231 \pm 0.017$  \\

\bottomrule
\end{tabular}
}
\caption{\textbf{Baseline Comparison}: SHADOW against six competing baselines.}
\label{tab:baselines}
\end{table*}

\noindent \textbf{Baseline Comparison } 
Table~\ref{tab:baselines} reports the performance of SHADOW compared to the 6 baselines. Overall, SHADOW consistently outperforms the competitors from the pursuer’s perspective, achieving the highest win rate of $P_{\text{win}} = 62\%$. The Periodic Communication strategy with $k=40$ achieves the second-highest win rate ($P_{\text{win}} = 57.6\%$), trailing SHADOW by 7.1\%. This difference is statistically significant according to a Mann-Whitney-U test, FDR-corrected $p=0.013$. Furthermore, the Periodic Communication with $k=40$ incurs a substantially higher probability of being shot ($P_{\text{shot}} = 41.6\%$) compared to SHADOW ($P_{\text{shot}} = 35\%$, a 15.8\% reduction, FDR-corrected $p=0.031$). Additionally, it results in longer episodes ($T_{\text{len}} = 320.1$ vs.\ $272.8$ for SHADOW, a 14.7\% decrease, FDR-corrected $p=0.034$) and a smaller average communication distance ($D_{\text{comm}} = 0.206$ vs.\ $0.240$ for SHADOW, a 14.2\% increase, FDR-corrected $p=1.85 \times 10^{-4}$), indicating that the agent tends to communicate at closer ranges—potentially increasing risk of beinh shot. Moreover, SHADOW induces a higher average steering cost on the evader ($\bar{S}_E = 0.231$) compared to all periodic strategies (FDR-corrected $p>3.81 \times 10^{-72}$). This behavior likely depends on SHADOW's higher communication frequency ($C_{ratio}=17.2\%$) which forces the evader into more evasive maneuvers. 


 All RL-based baselines, i.e., MultiHead PPO, P-DQN and HyAR, yield more conservative pursuer behaviors. These agents communicate significantly less than SHADOW ($C_{\text{ratio}} = 6.6\%$, $4.6\%$ and $0.2\%$, respectively), resulting in lower probability of being shot ($P_{\text{shot}} = 5.6\%$, $4.2\%$ and $0.2\%$, respectively). However, this comes at the cost of the win rate ($P_{\text{win}} = 27.2\%$, $39.6\%$, and $24.6\%$, respectively). In contrast, SHADOW successfully balances risk and reward, accepting moderate communication exposure ($P_{\text{shot}} = 35\%$) in exchange for significantly enhanced pursuit effectiveness.


\begin{figure}[t]
     \centering

     \subfloat[][]{\includegraphics[width=.23\textwidth]{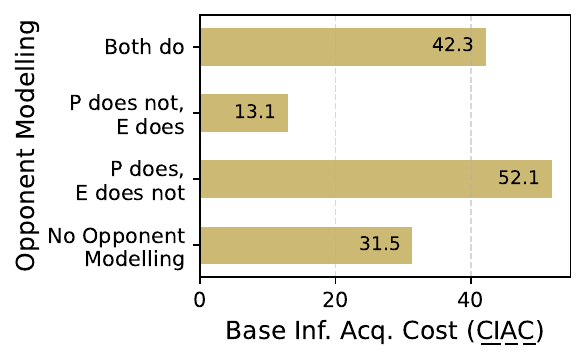}\label{fig:opponent_modeling_coi}}
     \subfloat[][]{\includegraphics[width=.23\textwidth]{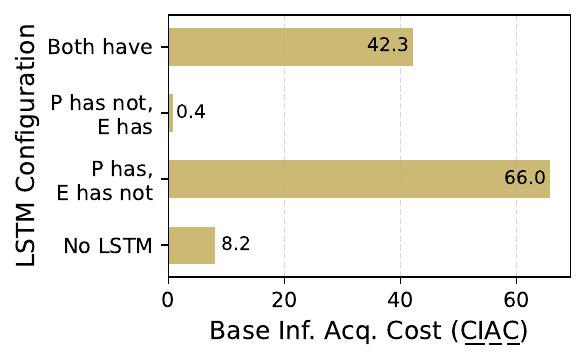}\label{fig:LSTM_coi}}
    
     \caption{\textbf{Ablation Study}: $\BIAC$ under different Opponent modeling (a) and LSTM (b) configurations.}
    
     \label{fig:LSTM_analysis_coi}
\end{figure}

\noindent \textbf{Ablation Study}
Figure~\ref{fig:opponent_modeling_coi} reports $\BIAC$ under four opponent modeling configurations. Equipping either agent with an opponent modeling module consistently yields a strategic advantage for that agent. When only the pursuer uses opponent modeling (\emph{P does, E does not}), $\BIAC$ reaches 52.1—the highest observed value, representing a substantial increase over the baseline of 31.5 when neither agent models its opponent (\emph{No Opponent Modeling}). Conversely, when only the evader uses opponent modeling (\emph{P does not, E does}), $\BIAC$ drops to 13.1, a notable decrease compared to 42.3 when both agents use opponent modeling (\emph{Both do}).

These results suggest that a pursuer that uses opponent modeling learns to communicate more strategically, accepting communication costs to gain higher returns. This holds regardless of the evader's configuration. The gain stems from two complementary factors: (i) less frequent but more effective communication, and (ii) increased likelihood of catching the evader. For instance, when the evader lacks opponent modeling, the pursuer’s $P_{\text{win}}$ rises from 47.2\% to 62\% (FDR-corrected $p=2.60 \times 10^{-6}$), while $C_{\text{ratio}}$ drops from 24\% to 15\% (FDR-corrected $p=4.18 \times 10^{-5}$), and $C_{\text{gap}}$ grows from 10.13 to 38.73 (FDR-corrected $p=4.98 \times 10^{-10}$). See Figures~\ref{fig:opponent_modeling}, \ref{fig:opponent_modeling_comm} in the Appendix for more details. 

A similar pattern emerges when applying opponent modeling to baselines powered by MultiHead PPO and P-DQN: in asymmetric settings, the agent with opponent modeling consistently outperforms its counterpart, while in the symmetric setting (where both agents use opponent modeling), the pursuer maintains a strategic edge. See Figure~\ref{fig:opponent_modeling_baselines} in the Appendix for additional results.


Finally, equipping either agent with an LSTM memory improves its performance. As shown in Figure \ref{fig:LSTM_coi}, equipping the pursuer with an LSTM yields a pursuer willing to pay more for information regarding the evader. For instance, when neither agent uses memory, $\BIAC = 8.2$. This value rises sharply to $\BIAC = 66.0$ when only the pursuer uses the LSTM. This improvement is primarily driven by a large increase in the pursuer’s success rate, with $P_\text{win}$ jumping from 11.2\% to 88.8\% (+87.3\%, FDR-corrected $p=3.73 \times 10^{-79}$). See Figures~\ref{fig:LSTM},\ref{fig:LSTM_comm} in the appendix for full details.

\begin{figure}[t]
     \centering
     \includegraphics[width=.33\textwidth]{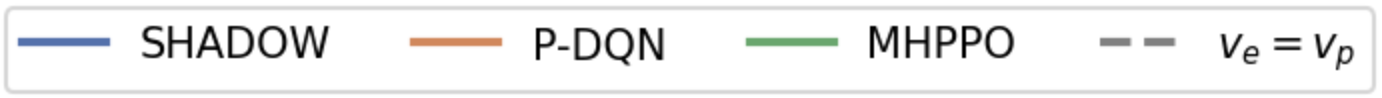}
     
     \subfloat[][]{\includegraphics[width=.22\textwidth]{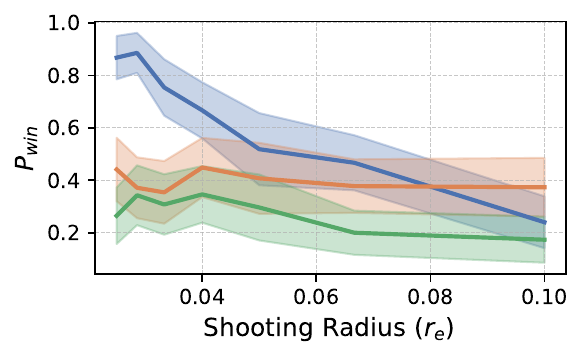}\label{fig:sa_lambd_Pwin}}
     \subfloat[][]{\includegraphics[width=.22\textwidth]{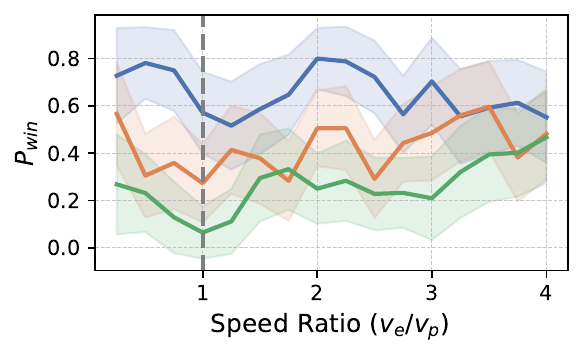}\label{fig:sa_speed_rario_Pwin}}

     \caption{\textbf{Sensitivity analysis}: Effect of the shooting radius $r_\eliminate$ (a) and the speed ratio $v_e / v_p$ (b) on pursuer win rate $P_\text{win}$.}
     \label{fig:sensitivity_analysis}
\end{figure}

\noindent \textbf{Sensitivity Analysis}
We examine how the shooting radius $r_\eliminate$ affects game's outcome metrics. Figure \ref{fig:sa_lambd_Pwin} shows the pursuer win rate $P_\text{win}$ as a function of $r_\eliminate$. SHADOW exhibits a clear decreasing trend: as communication becomes riskier (i.e., higher $r_\eliminate$), the pursuer's performance drops significantly. This is confirmed by regression analysis, where the (negative) effect size is substantial and highly significant (FDR-corrected $\beta = -28.3$, FDR-corrected $p = 1.45 \times 10^{-15}, R^2=0.144$). In contrast, P-DQN and MultiHead PPO do not show meaningful adaptation to changing risk: the same regression does not yield statistically significant results ($p > 0.10$). See Figure \ref{fig:sensitivity_analysis_app} in Appendix for visual trends.

We further investigate how $r_\eliminate$ modulates communication behavior. For SHADOW, an increase in $r_\eliminate$ leads to lower communication frequency ($C_\text{ratio}$) and shorter communication gaps ($D_\text{comm}$) from the shooting radius $r_e$ (see Figure \ref{fig:sensitivity_analysis_app} in the appendix for visual trends). Specifically, when we fit a regression model to infer $C_\text{ratio}$ and $D_\text{comm}$, we obtain the following results: \emph{(i)} For $C_\text{ratio}$: $\beta = -0.886$, FDR-corrected $p = 0.010$, $R^2 = 0.027$. \emph{(ii)} For $D_\text{comm}$: $\beta = 1.60$, FDR-corrected $p = 4.82 \times 10^{-6}$, $R^2 = 0.074$. In contrast, P-DQN and MHPPO show no significant relationship between $r_\eliminate$ and either $C_\text{ratio}$ or $D_\text{comm}$ (all FDR-corrected $p > 0.59$). This indicates that their policies are insensitive to variations in communication risk and fail to optimize the communication-efficiency trade-off as SHADOW does. See Figure \ref{fig:sensitivity_analysis_app} in the appendix for further details.

Next, we assess how the relative speed of the evader and pursuer, expressed as the ratio $v_e / v_p$, affects the win rate $P_{\text{win}}$ for SHADOW, P-DQN, and MHPPO, as shown in Figure \ref{fig:sa_speed_rario_Pwin}. We do not observe any trend: across all models, the regression results indicate no significant relationship between $P_{\text{win}}$ and the speed ratio (FDR-corrected $p>0.2$). Despite this, we find that the speed ratio influences communication behavior: for SHADOW and MultiHead PPO, higher speed ratio (i.e., faster evaders) lead to increased $\BIAC$ ($\beta = 7.48$, $p = 0.018$, $R^2 = 0.372$ for SHADOW; $\beta = 20.44$, $p = 0.005$, $R^2 = 0.518$), indicating that slower pursuers increasingly rely on communication to compensate the speed difference (see Figure~\ref{fig:sa_speed_rario_coi} in Appendix).


Finally, we analyze the dynamics of the uncertainty $\sigma$ predicted by the opponent modeling module over time. We find that $\sigma$ increases systematically in the timesteps leading up to a communication event ($q_p=1$), while no such trend is observed before generic non-communication actions. This suggests that the pursuer monitors the reliability of its internal predictions and tends to initiate communication when uncertainty accumulates. Full analysis and statistical results are provided in Appendix.



\begin{figure}[t]
    \centering
    \includegraphics[width=.85\linewidth]{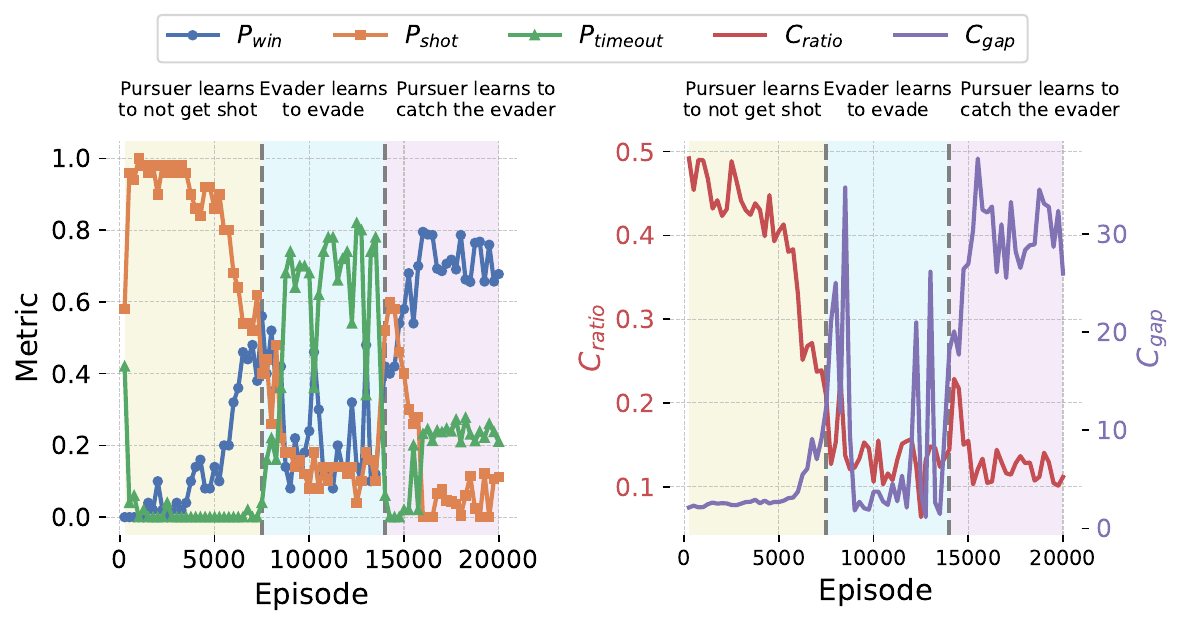}
    \caption{\textbf{Training dynamics}: Outcome and communication metrics over training episodes. Shaded regions corresponding to distinct learning phases.}
    \label{fig:training_dynamics}
\end{figure}

\noindent \textbf{Training Dynamics}
We analyze the evolution of the pursuer’s strategy during training by examining outcome and communication metrics across episodes. Figure \ref{fig:training_dynamics} reports the trends of $P_{\text{win}}$, $P_{\text{shot}}$, $P_{\text{timeout}}$, $C_{\text{ratio}}$, and $C_{\text{gap}}$ during training. The figure is divided into three shaded regions, each corresponding to a qualitatively different phase of training.

During an initial phase (up to episode 7,500), the pursuer rapidly learns to reduce the risk of being shot. This is reflected by a steep decline in $P_{\text{shot}}$ and $C_{\text{ratio}}$, indicating that the agent quickly recognizes the dangers of communication. 

During the intermediate phase (episodes 7,500 to 14,500), the agent enters a transient regime: the pursuer frequently times out as the evader learns an effective movement policy, i.e., its reward increases steeply (see Figure \ref{fig:rewards} in the appendix for further details). In this phase, the pursuer is actively experimenting with different communication strategies, i.e., while $C_{\text{ratio}}$ stabilizes around 15\%, $C_{\text{gap}}$ exhibits substantial fluctuations between 2 and 35. 

In the final phase (after episode 14,500), the pursuer's win rate improves and stabilizes above 60\%.  This improvement is \emph{not} accompanied by a change in $C_{\text{ratio}}$, which remains steady. Instead, the key adaptation occurs in $C_{\text{gap}}$, which stabilizes at a value of 30, indicating that the pursuer has learned to distribute evader's queries more strategically.


\section{Conclusion}

We generalized the PEEC framework proposed in \cite{huang2021pursuit} to model the strategic tension between information gathering and concealment in adversarial settings, and proposed \textbf{SHADOW}, an RL-based approach tailored to this challenge. Our formulation allows for asymmetric dynamics, non-zero-sum objectives, and realistic exposure risks where agents may be eliminated upon discovery. SHADOW pursuers learn to balance sensing and stealth through joint optimization of movement and communication, enhanced by opponent modeling. Empirical results show that SHADOW outperforms six baselines across a range of threat levels and environmental dynamics. Moreover, SHADOW pursuers adaptively modulate communication based on communication risk, leveraging predicted uncertainty to query more when uncertainty is high and remain silent when confidence is sufficient, leading to increasingly strategic timing over training.



\section{Acknowledgments}
This work is partly funded by ARO under grant W911NF2320240 and by Israel Science Foundation under grant 2544/24.

\bibliography{references}


\appendix

\begin{figure*}[t]
     \centering
     \includegraphics[width=.5\textwidth]{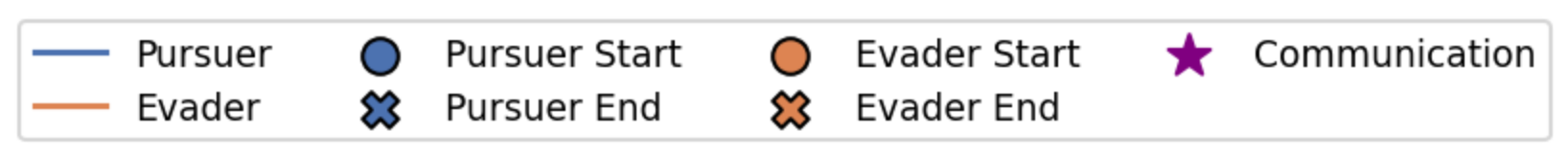}
     \subfloat[][]{\includegraphics[width=.24\textwidth]{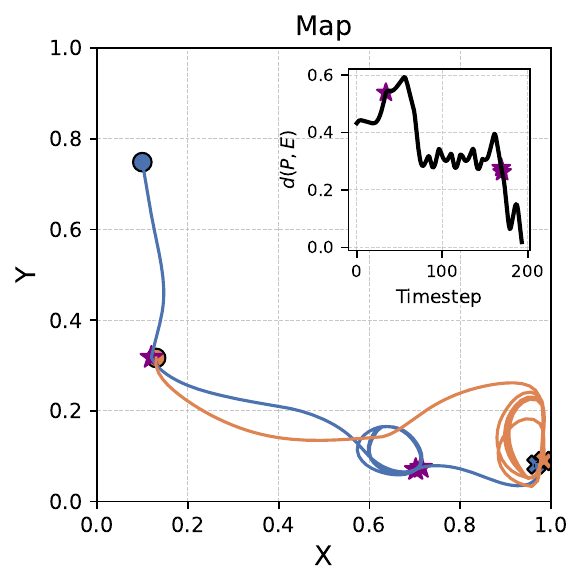}\label{fig:ex_1}}
     \subfloat[][]{\includegraphics[width=.24\textwidth]{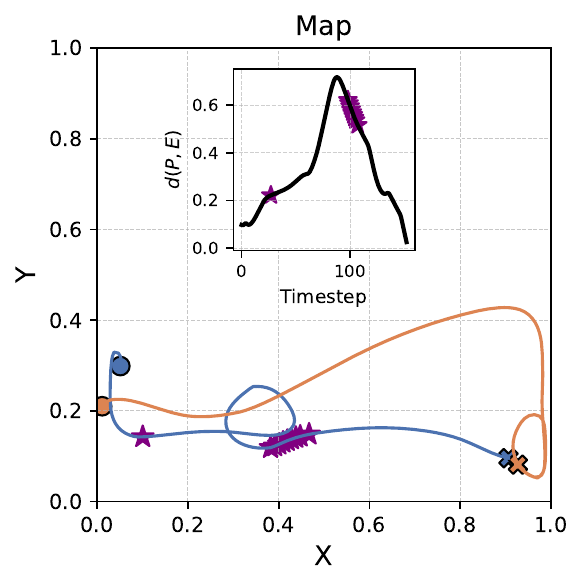}\label{fig:ex_2}}
     \subfloat[][]{\includegraphics[width=.24\textwidth]{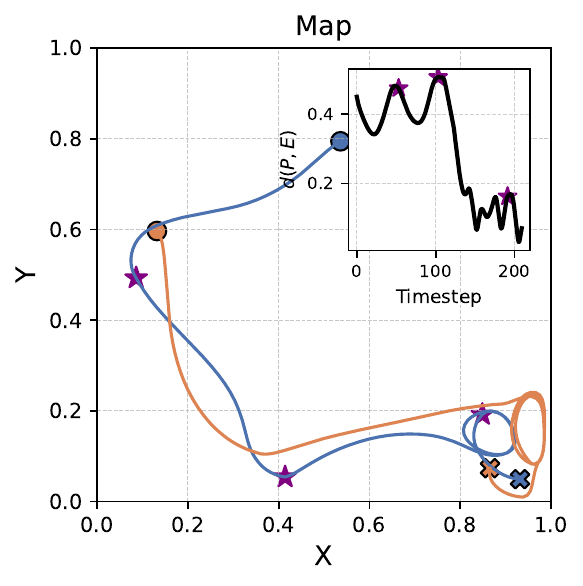}\label{fig:ex_3}}
     \subfloat[][]{\includegraphics[width=.24\textwidth]{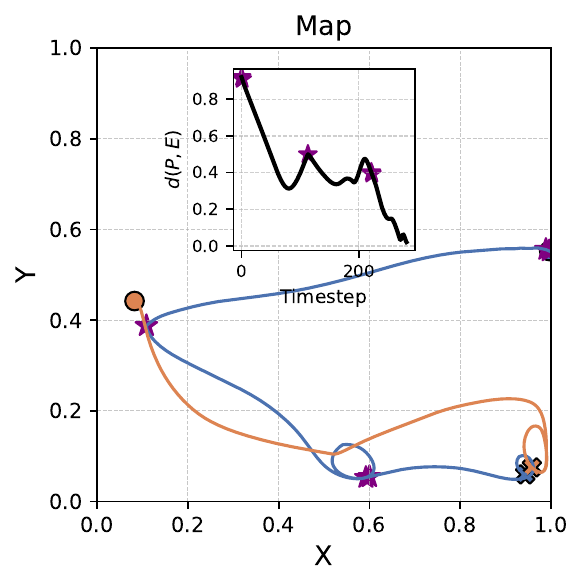}\label{fig:ex_4}}     
     \caption{ \textbf{Examples of pursuer-evader interactions in our PEEC game}: Each subplot illustrates a representative game, showing the trajectories of the pursuer and evader over the 2D map. The inset plot in each subplot reports the pursuer–evader distance as a function of the timestep. 
     }
    
     \label{fig:examples}
\end{figure*}

\section{Examples}
\label{app:examples}

Figure~\ref{fig:examples} shows four representative episodes in which the pursuer successfully captures the evader. In all cases, both agents are controlled using the SHADOW framework. These examples are selected to highlight the diversity and adaptiveness of the learned strategies, particularly the non-trivial communication behavior of the pursuer.

A consistent observation across all examples is that the pursuer’s communication strategy is not periodic. 
Initially, the pursuer typically attempts to intercept the evader by exploring the surrounding region near its starting position. If this initial search proves unfruitful, the pursuer triggers its first communication request to obtain an updated position of the evader. We note that in most cases, the pursuer does not initiate communication at the beginning of the episode. This depends on the assumption that the initial position of the evader is known to the pursuer, and vice-versa. The only exception is the example in Figure \ref{fig:ex_4}, where the pursuer starts on the extreme edge of the map with a heading directed toward the boundary. Given that the environment does not support toroidal wrapping, this configuration likely prompted an early communication request to avoid an inefficient trajectory.

Furthermore, a recurring behavioral motif is the looping or circular movement of the pursuer around the inferred position of the evader. This pattern appears in all examples and reflects a strategy to maximize the chance of intercepting the evader while minimizing unnecessary communication. The looping behavior continues until the pursuer performs another communication or closes the distance sufficiently to trigger capture.

The distance dynamics between the agents, shown in the inset plots, further illustrate the non-linear nature of the pursuit process. In Examples~\ref{fig:ex_1} and~\ref{fig:ex_3}, the distance fluctuates due to temporary misjudgments in the evader's inferred position. In Example~\ref{fig:ex_2}, the distance initially increases as the evader escapes, but is followed by a sharp decrease once the pursuer commits to a sequence of closely spaced communications. This suggests that the pursuer inferred a containment opportunity, ultimately cornering the evader in the lower-right region of the map. Interestingly, Example~\ref{fig:ex_2} also demonstrates that multiple consecutive communication actions—although seemingly redundant—might be effective when coordinated with spatial awareness.

\section{Implementation Details}
\label{app:impl_dits}

To promote generalization and avoid overfitting to specific scenarios, we introduced randomization in the initial conditions of each training episode. In particular, the initial positions of the pursuer and the evader were sampled uniformly over the map. Moreover, two key environment parameters were randomized: the shooting radius $r_e \sim \mathcal{U}(1, 4)$, and the speed ratio between the evader and the pursuer $v_e / v_p \sim \mathcal{U}(0.1, 4)$. We assume that both agents share the same maximum lateral acceleration, ensuring symmetry in maneuverability.

To monitor performance during training, we conducted validation every 250 episodes by running the current policy over 50 independent simulations. These intermediate evaluations provide insight into learning dynamics and are used to track policy convergence across experimental conditions.

All experiments were conducted on a high-performance computing workstation equipped with an Intel 10th Gen i9-10980XE processor, 256~GB of RAM, and an NVIDIA RTX A6000 GPU with 48~GB of dedicated memory. The simulation environment and learning algorithms were implemented in \texttt{Python 3.10} using \texttt{PyTorch 2.3}. To ensure full reproducibility of our results, we report in Table \ref{tab:hyperparams} the values of each hyperparameter of SHADOW and the environment.

\begin{table}[t]
\centering
\resizebox{\linewidth}{!}{
\begin{tabular}{crc} \toprule
Category & Parameter & Value  \\ \midrule

\multirow{7}{*}{Environment}  
  &  Maximum Mission Time (s), $\Horizon$ & $1000$ \\
  &  Map Size (km $\times$ km), $M=[0,W]\times [0,H] $ & $W=H=1$  \\
  &  Reference Speed (knot), $v_{\text{ref}}$ & $15$ \\
  &  Speed Ratio, $v_e / v_p$ & $ v_e / v_p \sim \mathcal{U}(0.1, 4)$ \\
  &  Agents' Acceleration (rad$/s$), $U_p,u_e$ & $0.9\pi$  \\
  &  Catching Radius (m), $r_\catch$ & $25$  \\
  &  Shooting Radius (m), $r_\eliminate$ & $r_\eliminate \sim \mathcal{U}(25, 100)$ \\ 
\midrule
\multirow{6}{*}{Reward} 
  &  Time Penalty, $\alpha_p^\Time,\alpha_e^\Time$ & 0.5, -0.5  \\
  &  Communication Penalty, $\alpha_p^\Query$ & 0 \\  
  &  Hit Boundary Penalty, $\alpha_p^\Boundary=\alpha_e^\Boundary$ & 10 \\
  &  Lateral Acceleration Penalty, $\alpha_p^\Acceleration=\alpha_e^\Acceleration$ & 0.5  \\
  & Terminal Reward - Shooting Penalty & 100  \\
  & Terminal Reward - Catch Bonus & 1000  \\
\midrule
\multirow{6}{*}{Sequential TD3} 
  & Soft Update Rate      & 0.005 \\  
  & Policy Noise         & 0.2    \\  
  & Noise Clip          & 0.5      \\  
  & Policy Delay        & 5       \\  
  & Exploration Noise     & 0.1   \\
\midrule
\multirow{2}{*}{Sequential PPO} 
  & Policy Clip        &  0.2     \\ 
  & $\lambda$ for Generalized Advantage Estimation         & 0.95     \\  
\midrule
\multirow{8}{*}{Shared} 
  &  No. Training Steps         & $10^7$ \\
  &  Evaluation Frequency       & $250$ \\
  &  No. Evaluation Episodes    & $50$   \\
  &  Replay buffer size         & $10^6$  \\
  &  Batch size                 & $32$  \\
  &  Hidden dimension           & $256$ \\

  &  Learning rate, $\eta$      & $3e^{-4}$  \\
  &  Discount factor, $\gamma$  & $0.99$  \\
\bottomrule
\end{tabular}
}
\caption{\textbf{Environment and model hyperparameters}.}
\label{tab:hyperparams}
\end{table}

\section{Metrics}
\noindent \textbf{End-State Outcomes} $P_{\text{timeout}} + P_{\text{win}} + P_{\text{shot}} = 1$ denote the fraction of the $N$ evaluation episodes in which the pursuer succeeds, is eliminated, or runs out of time (i.e., the evader escapes), respectively.



\noindent \textbf{Communication Strategy}
The average fraction of timesteps in which the pursuer communicates is defined as:
\begin{equation*}
C_{\text{ratio}}=\avg{i=1}{N}{\avg{t=1}{\Horizon_f^{(i)}}{q_t^{(i)}}},
\end{equation*}
where $\avg{i}{N}{X_i}$ denotes an average $N^{-1} (X_1+\ldots+X_N)$.

Let  $\{t_j^{(i)}\}_{j=1}^{N^{(i)}_\Query}$ be the ordered timesteps when communication occurs in episode $i$.
The average number of timesteps between consecutive communications is:
\begin{equation*}
C_{\text{gap}} = \avg{i=1}{N}{\avg{j=2}{N^{(i)}_\Query}{(t_j^{(i)} - t_{j-1}^{(i)})}}
\end{equation*}

The average distance between agents at the last communication is $D_{\text{comm}}= \frac{1}{N} \sum_{i=1}^N r^{(i)}(t_{M_i}^{(i)})$.

\noindent \textbf{Behavioral Efficiency} We assess the physical effort and timing of each episode. The average episode duration, in timesteps, is $T_{\text{len}} = \avg{i=1}{N}{\Horizon_f^{(i)}}$.

The average steering cost per timestep for the pursuer and evader are defined as $\bar{S}_P=\avg{i}{}{\avg{t}{}{u_p^{(i)}(t)}}$ and evader $\bar{S}_E=\avg{i}{}{\avg{t}{}{u_e^{(i)}(t)}}$, respectively.



\noindent \textbf{Information Acquisition Cost} We assess the lower bound on the cost of information acquisition as the average pursuer payoff over the average number of queries:
$$ \BIAC = \frac{\avg{i}{}{P_p^{(i)}}}{\avg{i}{}{N_\Query^{(i)}}} $$

\section{Baselines}

We examine whether SHADOW can outperform three heuristic and two learning-based strategies in terms of effectiveness and efficiency. Specifically, we compare SHADOW against six baselines:

\begin{enumerate}[label=\roman*.]
    \item \emph{No Communication}: the pursuer never communicates.
    \item \emph{Random Communication}: the pursuer uses a distance-dependent probability function to decide when to query the evader's state. It computes the current distance between the pursuer and the last observed evader $d(P,E)$, then estimates the probability of getting shot $p_{\text{shot}}$. The probability of communication is then set to $p_{\text{comm}}=1-p_{\text{shot}}$. 
    \item \emph{Periodic Communication}: the pursuer communicates periodically, each $k$ timesteps.
    Notably, this strategy has proven to be theoretically optimal in similar partially observable control settings~\cite{huang2022pursuitevasiondifferentialgamestrategic}, where shooting was not taken into account, and thus serve as a meaningful non-learning benchmark.
    \item \emph{MultiHead PPO} \cite{flet2019merl}: the pursuer leverages a multi-headed actor with PPO to jointly learn the communication and movement policies. 
    \item \emph{P-DQN} \cite{xiong2018parametrized}: the pursuer leverages a Parametrized Deep Q-Network to jointly learn the query and navigation policies. Notably, this baseline is specifically designed for hybrid action spaces, as in our environment the pursuer has to take a discrete action (query $q_p$) and a continuous action (lateral acceleration $u_p$).
    \item \emph{HyAR} \cite{DBLP:journals/corr/abs-2109-05490}: the pursuer uses discrete action (query action $q_p$) embeddings to capture high-level choices and a conditional variational autoencoder to generate continuous actions (lateral acceleration $u_p$) conditioned on these choices.
\end{enumerate}

We adapted publicly available and widely adopted implementations of Multihead PPO~\cite{pytorchrl} and Parameterized DQN (P-DQN)~\cite{bester2019mpdqn}. We developed our own implementation of HyAR since we did not find any official code-base for this approach. These models were adapted to operate under the same observation and action spaces as SHADOW, and trained under equivalent conditions to ensure fair comparison.

\section{Ablation Study}
\label{app:ablation_study}
\subsection{Opponent Modeling}
\begin{figure}[t]
     \centering

     
     \includegraphics[width=.9\linewidth]{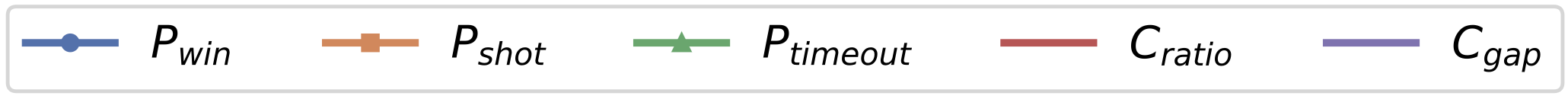}
     
     \subfloat[][]{\includegraphics[width=.25\textwidth]{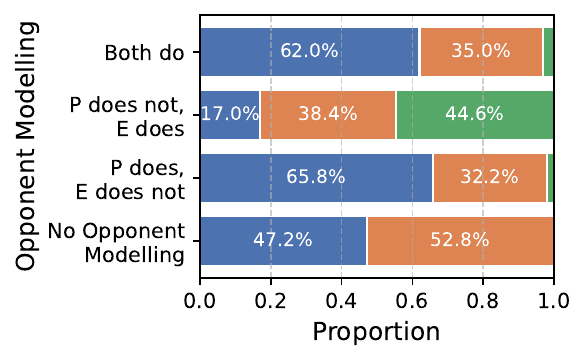}\label{fig:opponent_modeling}}     
     \subfloat[][]{\includegraphics[width=.25\textwidth]{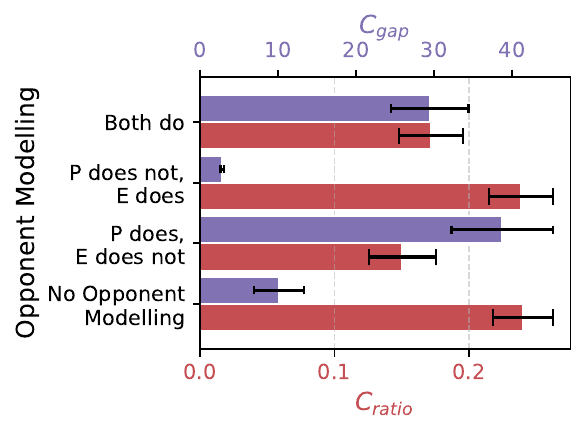}\label{fig:opponent_modeling_comm}} 

     \subfloat[][]{\includegraphics[width=.25\textwidth]{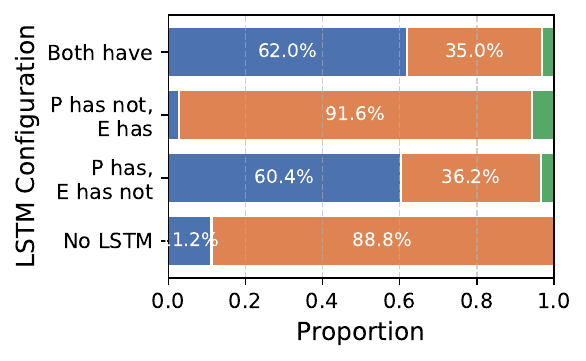}\label{fig:LSTM}}
     \subfloat[][]{\includegraphics[width=.25\textwidth]{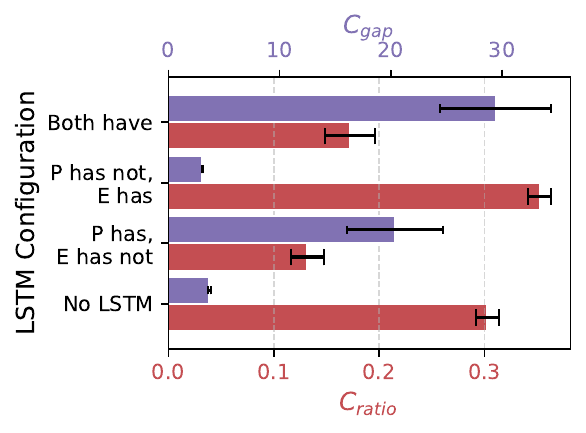}\label{fig:LSTM_comm}}

     \caption{\textbf{Ablation Study}: Game outcome and communication-related metrics for different opponent modeling configurations (a-b) and LSTM configurations (c-d).}
    
     \label{fig:}
\end{figure}

Figures \ref{fig:opponent_modeling} shows outcome-related metrics under four configurations of opponent modeling. We observe that equipping any agent with an opponent modeling module consistently provides a strategic advantage to that agent. When the pursuer is the sole user of opponent modeling (\emph{P does, E does not}), it achieves the highest win rate ($P_\text{win} = 0.658$), while the evader struggles to evade capture, resulting in the lowest timeout ($P_\text{timeout} = 0.02$) and shot ($P_\text{shot} = 0.322$) rates. In contrast, when only the evader employs opponent modeling (\emph{P does not, E does}), its ability to avoid capture increases significantly, reflected in a high timeout rate ($P_\text{timeout} = 0.446$) and a markedly reduced pursuer win rate ($P_\text{win} = 0.17$). 

Interestingly, in the symmetric setting, where neither agents utilize opponent modeling (\emph{No Opponent modeling}), the game is balanced, with near-equal win and shot rates ($P_\text{win} = 0.472$ and $P_\text{shot} = 0.528$, respectively), indicating neither agent holds a significant predictive advantage. Conversely, when both agents utilize opponent modeling (\emph{Both do}), the game  turns in favor of the pursuer ($P_\text{win} = 0.620$). 


To explain this difference, we investigate how opponent modeling influences the pursuer's communication strategy. We find in Figure \ref{fig:opponent_modeling_comm} that the opponent modeling module reduces reliance on frequent updates, enabling the pursuer to act more autonomously and efficiently. For example, in configuration \emph{P does, E does not}, we observe the lowest communication ratio ($C_\text{ratio} = 0.15$) and the largest average gap between communications ($C_\text{gap} = 38.73$). By contrast, the baseline configuration \emph{No Opponent modeling} yields a communication ratio $C_\text{ratio} = 0.24$ (37.5\% increase w.r.t. 0.15 obtained for the \emph{P does, E does not} configuration) and a average gap between communications $C_\text{gap} = 10.13$ (73.8\% decrease w.r.t. 48.73 for the \emph{P does, E does not} configuration). 



\begin{figure}[t]
     \centering

     \includegraphics[width=.25\textwidth]{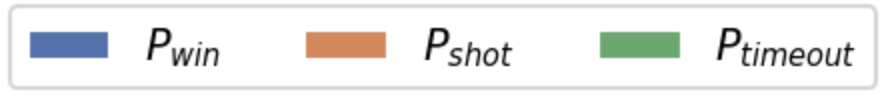}
     
     \subfloat[][MultiHead PPO]{\includegraphics[width=.25\textwidth]{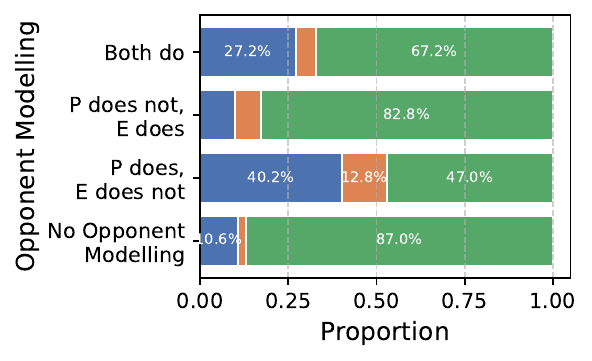}\label{fig:MHPPO_opponent_modeling}}
     \subfloat[][P-DQN]{\includegraphics[width=.25\textwidth]{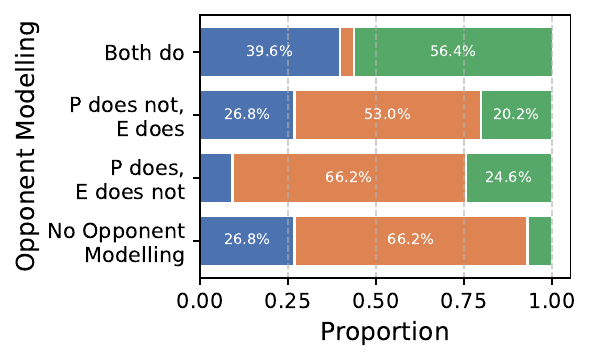}\label{fig:PDQN_opponent_modeling}}
 

     \caption{\textbf{Ablation Study}: Outcome-related metrics ($P_\text{win}$, $P_\text{shot}$, $P_\text{timeout}$) for the four configurations of opponent modeling for MultiHead PPO (a) and P-DQN (b) baselines.
     }
    
     \label{fig:opponent_modeling_baselines}
\end{figure}

Figure~\ref{fig:opponent_modeling_baselines} extends our ablation study by examining the impact of the opponent modeling module on two alternative baseline architectures: MultiHead PPO and P-DQN. While these methods differ in their absolute performance, the overarching trends observed with SHADOW remain consistent. For example, for MultiHead PPO in asymmetric settings, the agent equipped with opponent modeling consistently gains a strategic advantage over its counterpart.

In addition, when both agents employ opponent modeling, the pursuer improves its win rate, similar to the behavior observed in SHADOW. This trend is particularly evident in the case of P-DQN, where the pursuer’s win rate increases from 26.8\% (without opponent modeling) to 39.6\% (with opponent modeling enabled for both agents). These findings reinforce the conclusion that, irrespective of the evader’s configuration, incorporating opponent modeling is beneficial for the pursuer.

\subsection{LSTM-based Configurations}

Next, we investigate the role of the LSTM-based sequence encoder used to learn the query decision policy and navigation policy. Figure \ref{fig:LSTM} shows outcome-related metrics under the four LSTM configurations. The results indicate a substantial impact of temporal abstraction on the pursuer's performance. For example, when neither agent uses an LSTM, the pursuer rarely succeeds ($P_{\text{win}} = 0.112$), being always shot down before interception ($P_{\text{shot}} = 0.888$). In contrast, when the pursuer is equipped with an LSTM and the evader is not, the pursuer’s win rate increases dramatically to 0.604—mirroring the effect of the opponent modeling module. 

The LSTM also benefits the evader: in the asymmetric setup where only the evader uses an LSTM, the pursuer’s evasion rate, i.e., $P_{\text{timeout}}$, increases from 0 to 0.056. Finally, the full symmetric configuration (both agents using LSTM) yields the highest performance overall ($P_{\text{win}} = 0.620$), suggesting that access to temporal patterns benefits both agents, but still results in a net advantage for the pursuer.

Beyond win/loss outcomes, Figure \ref{fig:LSTM_comm} shows that the LSTM encoder has the same effect on the communication policy learnt by the pursuer, i.e., it allows the agent to communicate less frequently (lower $C_\text{ratio}$ and larger $C_\text{gap}$). This is likely because this sequential model acts like a memory to remember recent communications\footnote{While we include the elapsed time since last communication in the state of the game, the LSTM-based memory may also represent previous communications and their effect on agent's dynamics. }.

\section{Sensitivity Analysis}
\label{app:sa}

\subsection{Shooting Radius}

\begin{figure}[t]
     \centering
     \includegraphics[width=.4\textwidth]{figures/legend_sensitivity_analysis.png}
     \subfloat[][]{\includegraphics[width=.25\textwidth]{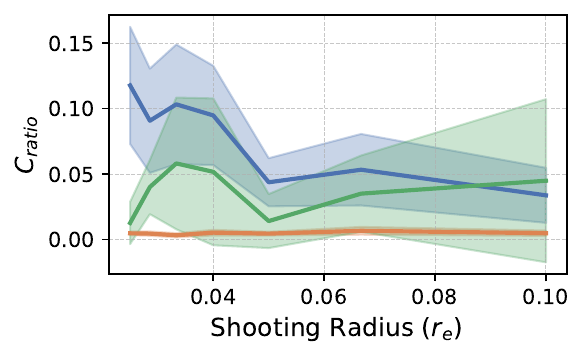}\label{fig:sa_lambd}}
     \subfloat[][]{\includegraphics[width=.25\textwidth]{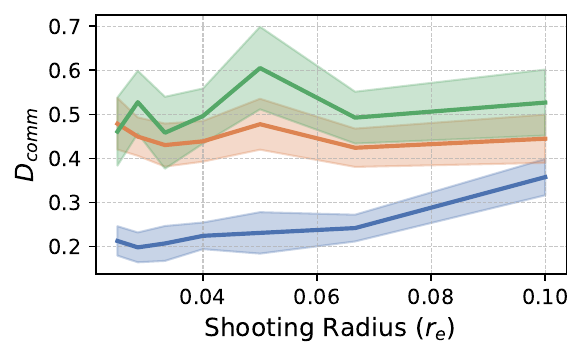}\label{fig:sa_speed_ratio}}
     
     \caption{\textbf{Sensitivity analysis}: Effect of the shooting radius $r_e$ on the communication ratio $C_{\text{ratio}}$ and the average distance at last communication $D_\text{comm}$.}
    
     \label{fig:sensitivity_analysis_app}
\end{figure}

Figure~\ref{fig:sensitivity_analysis_app} shows the effect of the shooting radius $r_\eliminate$ on the communication ratio $C_{\text{ratio}}$ and the average distance at last communication $D_{\text{comm}}$ for SHADOW, P-DQN and MultiHead PPO. 

As discussed in the main paper, SHADOW exhibits clear and significant trends: as $r_\eliminate$ increases, communication involves more risk, and so the pursuer communicates less frequently (lower $C_{\text{ratio}}$) and with larger intervals between communications (larger $D_{\text{comm}}$). These behavioral adaptations reflect a risk-aware communication policy that dynamically adjusts to the cost of information disclosure.

In contrast, both MHPPO and P-DQN do not show meaningful variation: their $C_{\text{ratio}}$ and $D_{\text{comm}}$ remain stable as $r_\eliminate$ varies, reinforcing the conclusion that these methods are insensitive to changes in communication risk. 

Next, we investigate how the shooting radius $r_\eliminate$ influences the information acquisition cost directly ($\BIAC$). Figure \ref{fig:sa_lambd_coi} shows $\BIAC$ as a function of $r_\eliminate$ for SHADOW, P-DQN, and MultiHead PPO. Across all values of $r_\eliminate$, pursuers controlled by P-DQN and MultiHead PPO exhibit higher $\BIAC$, which depends on their lower communication frequency (see Table \ref{tab:baselines}). Notably, only SHADOW displays a clear decreasing trend: as $r_\eliminate$ increases, $\BIAC$ decreases. In other words, the greater the threat of being shot, the more selectively the pursuer communicates. This behavioral adaptation is supported by a regression analysis using $r_\eliminate$ as the independent variable ($\beta = -330.9$, FDR-corrected $p = 0.024$, $R^2 = 0.669$). In contrast, P-DQN and MultiHead PPO show no significant sensitivity to risk level, with FDR-corrected $p > 0.48$ in both cases.

\subsection{Speed Ratio}

\begin{figure}[t]
     \centering
     \includegraphics[width=.4\textwidth]{figures/legend_sensitivity_analysis.png}
     
     \subfloat[][]{\includegraphics[width=.25\textwidth]{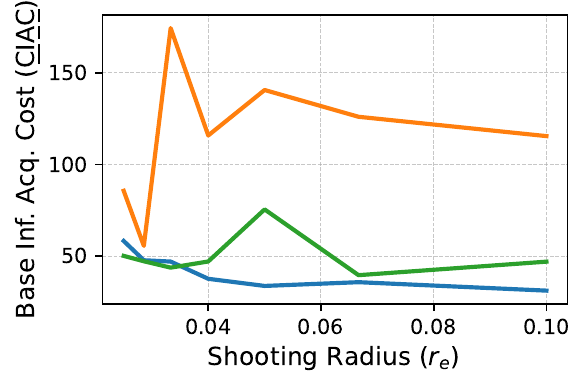}\label{fig:sa_lambd_coi}} \subfloat[][]{\includegraphics[width=.25\textwidth]{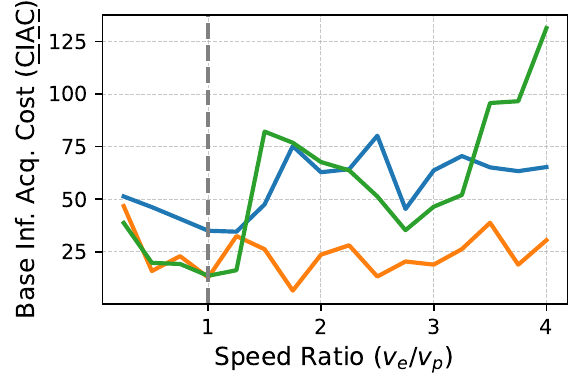}\label{fig:sa_speed_rario_coi}}
     
     \caption{\textbf{Sensitivity analysis}: Effect of the shooting radius $r_\eliminate$ (a) and the speed ratio $v_e / v_p$ (b) on $\BIAC$.}
     \label{fig:sensitivity_analysis_appendix}
\end{figure}

We examine the impact of the evader-to-pursuer speed ratio $v_e / v_p$ on $CoI$. As shown in Figure \ref{fig:sa_speed_rario_coi}, SHADOW and MultiHead PPO both exhibit a positive correlation, indicating that as the evader becomes relatively faster, $\BIAC$ increases ($\beta = 7.48$, FDR-corrected $p = 0.018$, $R^2 = 0.372$ for SHADOW; $\beta = 20.44$, FDR-corrected $p = 0.005$, $R^2 = 0.518$ for MultiHead PPO). This suggests that greater speed disparities in favor of the evader compel the pursuer to rely more heavily on communication. A likely explanation is that faster evaders require more adaptive and coordinated responses, increasing the value of communication. In contrast, P-DQN shows no statistically significant trend ($p > 0.64$), suggesting limited behavioral responsiveness to changing evader dynamics.

\subsection{Uncertainty of the Opponent Position}

\begin{figure}[t]
     \centering

     \subfloat[][]{\includegraphics[width=.25\textwidth]{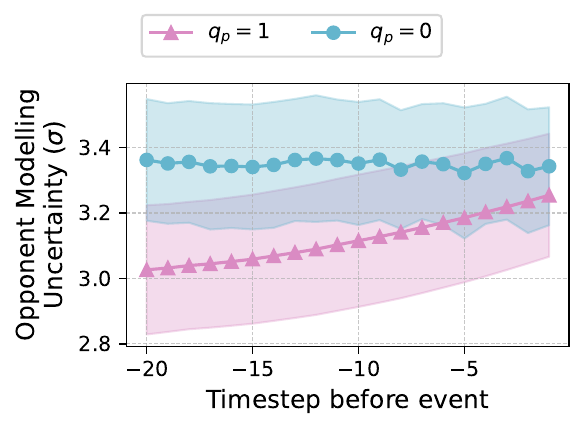}\label{fig:sigma}}
     \subfloat[][]{\includegraphics[width=.25\textwidth]{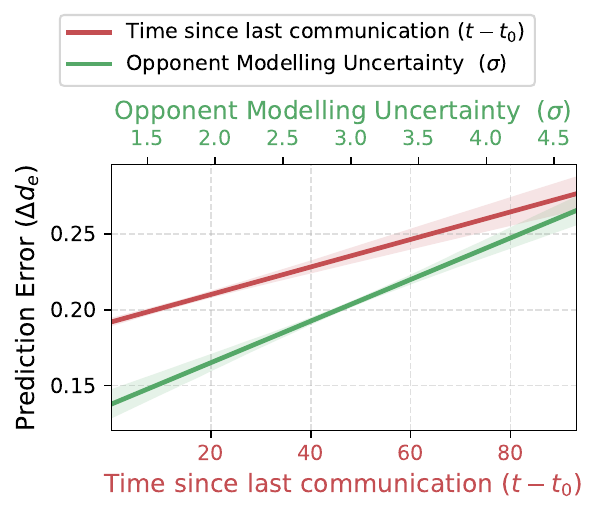}\label{fig:d_err_vs_comm_sigma}}

     \caption{\textbf{Sensitivity Analysis}: (a) Temporal evolution of predicted uncertainty $\sigma$ in the 20 timesteps preceding two types of events: communication actions ($q_p = 1$) and non-communication actions ($q_p = 0$). (b)  Prediction error $\Delta d_e$ of the opponent modeling module, measured as the distance between the evader’s true and predicted positions. $\Delta d_e$ is plotted as a function of the predicted uncertainty $\sigma$ and the time since the last communication ($t-t_0$).
     }
    
     \label{fig:sa_uncertainty}
\end{figure}

We examine the dynamics of the uncertainty $\sigma$ predicted by the opponent modeling module over time. Our analysis focuses on two conditions: (i) the 20 timesteps leading up to a communication event ($q_p=1$), and (ii) the 20 timesteps preceding a generic non-communication event ($q_p=0$), which serves as a control group. Figure~\ref{fig:sigma} shows the temporal trends of $\sigma$ under both conditions. A linear regression identifies a significant upward trend in uncertainty prior to communication events ($\beta = 0.012$, FDR-corrected $p = 3.81 \times 10^{-16}$), while no significant trend is observed in the control condition ($p = 0.258$). These findings suggest that the pursuer tends to query the evader's state when uncertainty about its position accumulates, highlighting the role of the opponent modeling module in guiding information-seeking behavior.

We further analyze the relationship between the predicted uncertainty $\sigma$ and the time since the last communication ($t - t_0$). Both variables are intuitively related to the evader's position:  Indeed, longer intervals since the last observation are expected to increase uncertainty, and similarly, high $\sigma$ explicitly signals unreliable predictions from the oppponent modeling module. Our goal is to understand whether these two variables provide overlapping or complementary information. To investigate this, we measure the prediction error $\Delta d_e = \lVert \ss_e - \ss_e' \rVert_2$, defined as the distance between the evader’s true position $\ss_e$ and the predicted position $\ss_e'$. We then regress this error on both $\sigma$ and $t - t_0$. As shown in Figure~\ref{fig:d_err_vs_comm_sigma}, both variables are significantly and positively associated with prediction error ($\beta = 0.0219$, FDR-corrected $p = 2.91 \times 10^{-48}$ for $\sigma$; $\beta = 0.0466$, FDR-corrected $p = 5.34 \times 10^{-36}$ for $t - t_0$). These results suggest that $\sigma$ and elapsed time since last communication capture different aspects of uncertainty, and imply that the pursuer may benefit from using both signals to guide its movement and communication strategies.

\section{Training Dynamics}
\label{app:train_dynamics}
\begin{figure}[t]
     \centering
     \subfloat[][]{\includegraphics[width=.25\textwidth]{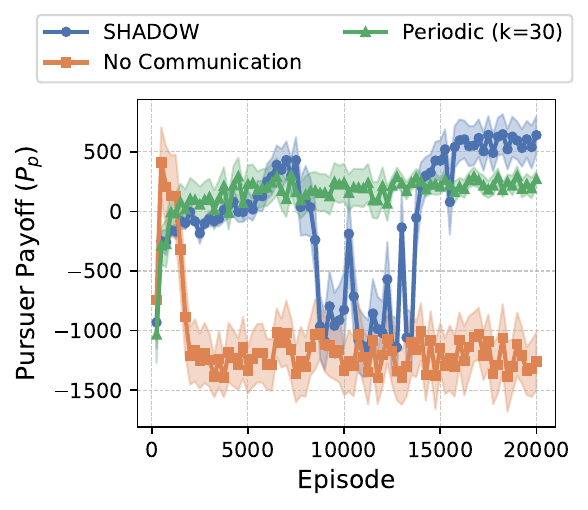}\label{fig:rew_p}}
     \subfloat[][]{\includegraphics[width=.25\textwidth]{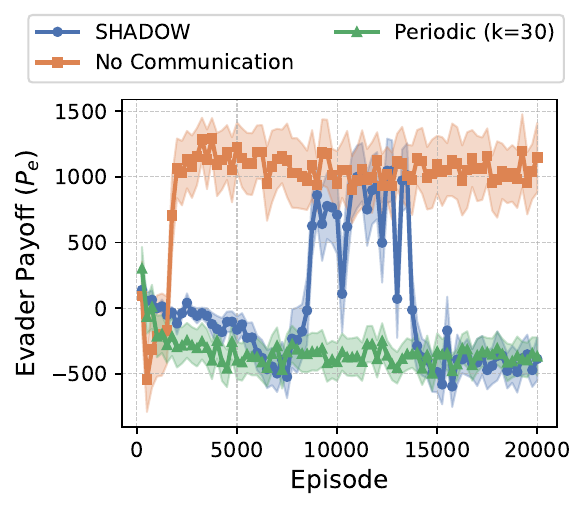}\label{fig:rew_e}}
     
     \caption{\textbf{Training Dynamics}: Cumulative reward obtained by the pursuer (a) and the evader (b) across training episodes under three strategies: SHADOW, periodic communication, and no communication (baseline).}
    
     \label{fig:rewards}
\end{figure}

Figure~\ref{fig:rewards} shows the cumulative rewards achieved by the pursuer and the evader throughout the training process. We compare three settings: (i) our proposed SHADOW method, (ii) a baseline with periodic communication (fixed interval $k = 30$), and (iii) a no-communication baseline.

In the No Communication scenario, the pursuer is unable to achieve any meaningful success. Its reward rapidly converges to values below 1,000 by episode 2,500, indicating that it consistently fails to intercept the evader. Meanwhile, the evader accumulates high rewards, reflecting its repeated success in escaping. However, this case is degenerate: due to the complete absence of communication, the pursuer is effectively blind, and the evader's high reward is not the result of a refined evasive strategy but rather a static failure of the learning process to progress.

In the periodic communication setting, training converges quickly (by episode 5,000). The pursuer achieves a moderate reward plateau of approximately 200, indicating that fixed communication intervals provide enough situational awareness to enable occasional successes, albeit without strategic adaptability.

In contrast, SHADOW exhibits a slower but more structured learning trajectory, consistent with the multi-stage training dynamics described in the main paper: Initially, the pursuer prioritizes minimizing the risk of being shot; only in later episodes does it fine-tune its communication timing to improve interception performance. This results in a longer convergence period (approximately 15,000 episodes), but ultimately leads to a higher average reward (approximately 500) than both baselines. 

\section{Comprehensive Related Work}
Pursuit-Evasion Differential Games (PEGs)~\cite{weintraub2020introduction} form a canonical framework for adversarial motion planning and have been studied extensively in the robotics literature, with applications that span from autonomous missile interception~\cite{xi2022nonlinear} and air-combat maneuvering~\cite{kothari2014cooperative} to self-driving cars~\cite{zhang2021safe} and multi-robot security patrol~\cite{shishika2020cooperative}. A growing body of work relaxes the idealized assumption of perfect information. We group prior works on PEGs with imperfect information into three categories that we scan below.

\paragraph{Exogenous visibility limits.} In this category, access to state information is constrained by external environmental factors. \cite{bhattacharya2010existence} proved existence of Nash equilibria despite obstacle-induced blind spots and line-of-sight constraints. Discrete-time counterparts show that scheduled noisy measurements qualitatively change optimal strategies~\cite{bopardikar2008discrete}, while incomplete-information variants highlight the need for randomized evader policies when the pursuer’s observations are noisy~\cite{hexner2019pursuit}. In all of these works, the access to the state is limited by the environment, which may return a noisy observation, the local observation up to a given radius, or the state at predetermined set of discrete time steps. In particular, players cannot decide when to acquire additional information.

\paragraph{Internal sensing cost.} Another line of work endows agents with sense or communicate actions that incur purely internal costs. In this setting, players can strategically decide to acquire information, but such queries come at a certain cost that can model e.g., battery capacity~\cite{maity2023efficient}, limited bandwidth~\cite{aleem2015self}, or link-establishment cost~\cite{maity2024optimal}. The cost can be either integrated as part of the player's payoff, imposing a soft constraint, or as a hard constraint, such as a budget that limits the number of state queries. To this end, \cite{maity2017linear} considered a switched-link linear-quadratic (LQ) differential game, were a fixed cost per measurement is introduced. Some extensions cover discrete-event formulations~\cite{maity2016strategies,maity2017stochastic}, finite-budget intermittent sensing~\cite{maity2023optimal}, and asymmetric settings where only the disadvantaged player can request costly observations~\cite{aggarwal2024linear,aggarwal2024best}. A related study features a static remote sensor whose limited transmissions—or deliberate silence—shape the game’s information flow~\cite{maity2023efficient}. In general, these papers yield limited sensing policies and quantify the performance loss under scarce measurements.

\paragraph{Implicit exposure cost.} As mentioned earlier, only~\cite{huang2021pursuit} explicitly consider information acquisition at the expense of information disclosure. While pioneering, their work assumed a relatively structured PEG that allows closed-form solutions and theoretical analysis on the one hand, but is limited in its applications in the other hand.

In this work, we alleviate assumptions taken in~\cite{huang2021pursuit}, and resort to Deep Reinforcement Learning (DRL) techniques in order to achieve a robust methodology for PEECs that applies to a broader and more realistic class of PEECs. In particular, we do not restrict ourselves to LQG games, and consider non-holonomic, non-linear dynamics and non-quadratic payoffs. Moreover, we do not have a global system payoff to be minimized by the evader and maximized by the pursuer. Instead, each player has its own payoff that it aims to minimize, and in turn the game may not be zero sum. Even in this broader setting, we empirically observe that without explicit communication cost, the pursuer should communicate frequently, even when it is slower than the evader and less maneuverable to some extent. We therefore strengthen the evader and allow it to eliminate a pursuer that discloses its location by communicating, with some probability that is proportional to the distance between them.

\paragraph{Reinforcement-Learning Solutions to PEGs (Without Exposure Cost).} Recently, an increasing number of works utilizes DRL tools to solve non-LQ PEGs with great success, yet does consider binding sensing to revealing information. Examples include multi-agent RL for underwater target-hunting with communication delay~\cite{wei2022underwater,wei2023differential}; deep-deterministic-policy-gradient guidance against maneuvering evaders~\cite{xi2024deep}; microswimmer pursuit–evasion at low Reynolds number~\cite{borra2022reinforcement}; decentralized multi-agent RL that scales to physical quadrotor pursuits~\cite{de2021decentralized}; and more~\cite{xu2022pursuit,hu2024transfer,wang2025hierarchical}. These studies demonstrate that model-free methods can address nonlinear dynamics, partial observability, and multi-agent coordination, yet none integrates an exposure/detection effect.

\section{Omitted Proofs}

\CiacPositive*

\begin{proof}
We first prove the existence of a maximum $\alpha_c^\Query$.

Since we assume a two-player zero-sum game, by the Min-Max Theorem~\cite{von1959mathematische}, the payoff in all Nash-equilibria is the same, and equals:
$$ V(\alpha_p^\Query):= \max_{\Omega_\NE[\alpha_p^\Query]}{\EE[P_p\mid \alpha_p^\Query]} = \max_{(u_p,q_p)}{\min_{u_e}{P_p(\langle (u_p,q_p); u_e \rangle \mid \alpha_p^\Query)}}$$. 

Since the space of pursuer and defender strategies is compact, it follows from the Maximum (Minimum) Theorem~\cite{berge1877topological} that $V(\alpha_p^\Query)$ is continuous. Moreover, $V$ is monotonic decreasing with $\alpha_p^\Query$. Indeed, let $\alpha_1 > \alpha_2$, and let $\langle (u_p^1, q_p^1); u_e^1 \rangle \in \Omega_\NE[\alpha_1]$. Then we have:
\begin{align*}
V(\alpha_2) &= \max_{(u_p^2,q_p^2)}{\min_{u_e^2}}{P_p(\ldots \mid \alpha_2)} \\
&\ge \min_{u_e^2}{P_p(\langle (u_p^1, q_p^1); u_e^2 \rangle \mid \alpha_2)} \\
&= P_p(\langle (u_p^1, q_p^1); u_e^1 \rangle \mid \alpha_2) \\
&\ge P_p(\langle (u_p^1, q_p^1); u_e^1 \rangle \mid \alpha_1) = V(\alpha_1),
\end{align*}
Where in the first inequality we use the same pursuer strategy for the lower communication penalty, the equality is then since the communication penalty is independent of the evader strategy and since $\langle (u_p^1, q_p^1); u_e^1 \rangle$ is in equilibrium, and the last inequality is due to $P_p$ being linearly decreasing with the communication penalty, when the strategy profile is fixed.

Finally, $\lim_{\alpha_p^\Query \rightarrow -\infty}{V(\alpha_p^\Query)} = \infty $. This is because when $\alpha_p^\Query$ is negative, the pursuer gets a reward of $|\alpha_p^\Query| \rightarrow \infty$ for a single query.

Therefore, the set $\Omega_V := \{\alpha_p^\Query \mid V(\alpha_p^\Query) \ge 0\} = V^{-1}([0, \infty))$ has a maximum, $\alpha_c^\Query$.

Next, let $\alpha_p^\Query<0$ and assume $r_e=0$. We will construct a pursuit strategy with a positive payoff. Let $d_0=\EE[D_0]>0$ be the expected distance between the pursuer and the evader at $t=0$. By Markov inequality, with probability at least 0.5, the initial distance is at least $d_0/2$. Let $t_0>0$ be the minimum time it takes the pursuer and evader to meet given initial distance is $\ge d_0/2$. For instance, in our game, $t_0=\frac{d_0/2-r_c}{(v_p+v_e)}$ as the players move at a constant speed. Finally, denote by $-P_{\max{}}$ the maximal negative payoff the pursuer can get. Consider the query strategy $q_p(\alpha_p^\Query)$ that queries the state every 
$${dt}_0 := \frac{t_0}{3\lceil P_{\max{}} / \alpha_p^\Query \rceil} $$
times starting with $t=0$, and let $u_p$ be arbitrary. Then the pursuer expected payoff equals:
\begin{align*}
\EE[P_p] =& \EE[P_p \mid D_0 > d_0/2] \Pr[D_0 > d_0/2] + \\
& \EE[P_p \mid D_0 \le d_0/2] \Pr[D_0 \le d_0/2] \ge \\
& (3-1) P_{\max{}} \cdot 0.5 - P_{\max{}} = 0,
\end{align*}
where we bound the second term by $-P_{\max{}}$, and for the first term, given $D_0>d_0/2$ (w.p. at least 0.5), the pursuer is ensured to perform at least $3P_{\max{}}/\alpha_p^\Query$ queries, and therefore receive a communication reward of $3P_{\max{}}$ overall.

Thus, for any negative communication penalty, there exist a maxi-min strategy for the pursuer yielding a non-negative expected payoff. Therefore, the supremum over $V^{-1}([0, \infty))$ must be non-negative, and so $\alpha_c^\Query \ge 0$.
\end{proof}

\BiacLB*

\begin{proof}
Recall for the case of a zero-sum game, we defined:
$$ V(\alpha_p^\Query):= \max_{(u_p,q_p)}{\min_{u_e}{P_p(\langle (u_p,q_p); u_e \rangle \mid \alpha_p^\Query)}},$$
and $\alpha_c^\Query$ is defined as the maximum communication penalty for which $V$ is non-negative. It is therefore sufficient to prove that $V(\underline{\alpha}_c^\Query) \ge 0$. Indeed:

\begin{align*}
V(\underline{\alpha}_c^\Query) &\ge \min_{u_e'}{\EE[P_p(\langle (u_p,q_p); u_e' \rangle \mid \underline{\alpha}_c^\Query)]} \\
&= \EE[P_p(\langle (u_p,q_p); u_e' \rangle \mid \underline{\alpha}_p^\Query)] \\
&= \EE[P_p(\langle (u_p,q_p); u_e' \rangle \mid 0)] - \underline{\alpha}_p^\Query \EE[N_p^\Query] \ge 0,
\end{align*}
where the first inequality is due to fixing a pursuer strategy, the following equality is since the evader is not penalized for pursuer communication, and the last inequality is by the definition of $\BIAC$.
\end{proof}

\end{document}